\pdfoutput=1
\documentclass[11pt]{article}

\usepackage{acl}

\usepackage{times}
\usepackage{latexsym}

\usepackage[T1]{fontenc}
\usepackage[utf8]{inputenc}

\usepackage{microtype}

\usepackage{inconsolata}

\usepackage{graphicx}

\usepackage{amsmath,amsfonts,bm}

\def\eqref#1{equation~\ref{#1}}
\def\1{\bm{1}}

\def\vx{{\bm{x}}}

\def\vz{{\bm{z}}}

\DeclareMathAlphabet{\mathsfit}{\encodingdefault}{\sfdefault}{m}{sl}
\SetMathAlphabet{\mathsfit}{bold}{\encodingdefault}{\sfdefault}{bx}{n}

\usepackage{url}
\usepackage{amsmath}
\usepackage{times}
\usepackage{framed}
\usepackage{ragged2e}
\usepackage{bm}
\usepackage{soul}
\usepackage{latexsym}
\usepackage{subfigure}
\usepackage{amsthm}
\usepackage{graphicx}
\usepackage{float}
\usepackage{longtable}
\usepackage{booktabs} 
\usepackage{multirow}
\usepackage{multicol}
\usepackage{amssymb}
\usepackage{amsmath}
\usepackage{dashbox}%
\usepackage{multirow}
\usepackage{textcomp}
\usepackage{tcolorbox}
\usepackage{makecell}
\usepackage{bbding}
\usepackage[ruled, vlined, linesnumbered]{algorithm2e}
\usepackage{mathtools}
\usepackage{adjustbox}
\usepackage[ruled,vlined]{algorithm2e}
\usepackage{color, soul}
\usepackage{pifont}
\usepackage{wrapfig}
\usepackage{xcolor}
\usepackage{array}
\usepackage{tabularx}
\newcolumntype{L}{>{\RaggedRight\hangafter=1\hangindent=0em}X}
\newcolumntype{P}[1]{>{\centering\arraybackslash}p{#1}}
\newcolumntype{M}[1]{>{\centering\arraybackslash}m{#1}}
\usepackage{cleveref}
\crefname{section}{§}{§§}
\Crefname{section}{§}{§§}
\crefname{figure}{Figure}{Figure}
\Crefname{figure}{Figure}{Figure}
\crefname{table}{Table}{Table}
\Crefname{table}{Table}{Table}
\definecolor{my_red}{RGB}{255,99,71}
\definecolor{my_green}{RGB}{50,205,50}
\definecolor{my_blue}{RGB}{65,105,225}
\definecolor{darkgreen}{rgb}{0, 0.70, 0}

\title{Pre-training Distillation for Large Language Models:\\ A Design Space Exploration}
\author{Hao Peng$^{1}$\footnotemark[2], Xin Lv$^2$, Yushi Bai$^{1}$\footnotemark[2], Zijun Yao$^{1}$\footnotemark[2], Jiajie Zhang$^{1}$\footnotemark[2], Lei Hou$^1$, Juanzi Li$^1$\\
$^{1}$Tsinghua University\quad 
$^{2}$Zhipu AI \\ 
\texttt{\{peng-h24\}@mails.tsinghua.edu.cn}}

\begin{document}
\maketitle
\renewcommand{\thefootnote}{\fnsymbol{footnote}}
\footnotetext[2]{Work is done when interned at Zhipu.AI}
\renewcommand*{\thefootnote}{\arabic{footnote}}

\begin{abstract}
Knowledge distillation (KD) aims to transfer knowledge from a large teacher model to a smaller student model. Previous work applying KD in the field of large language models (LLMs) typically focused on the post-training phase, where the student LLM learns directly from instructions and corresponding responses generated by the teacher model. 
In this paper, we extend KD to the pre-training phase of LLMs, named \textbf{pre-training distillation} (PD). We first conduct a preliminary experiment using GLM-4-9B as the teacher LLM to distill a 1.9B parameter student LLM, validating the effectiveness of PD. 
Considering the key impact factors of distillation, we systematically explore the design space of pre-training distillation across four aspects: logits processing, loss selection, scaling law, and \textit{offline} or \textit{online} logits. 
We conduct extensive experiments to explore the design space of pre-training distillation and find better configurations and interesting conclusions, such as larger student LLMs generally benefiting more from pre-training distillation, while a larger teacher LLM does not necessarily guarantee better results. 
We hope our exploration of the design space will inform future practices in pre-training distillation.

% Specifically, we use the GLM-4-9B and GLM-4-32B models as the teacher models and use the teacher model's output logits to train smaller models using negative log-likelihood loss. We find that distillation pretraining generally results in better performance compared to conventional pretraining. We further explore the best practice of key factors of distillation pretraining, including loss and logits selection. We also investigate the scaling law of distillation pretraining and find that a larger student model usually benefits more from distillation pretraining, while a larger teacher model does not necessarily guarantee better performance. Finally, we discuss the potential usage of distillation pretraining and directions for future work.

\end{abstract}

\section{Introduction}
Knowledge distillation (KD; \citealp{hinton2015distilling}) aims to distill the knowledge of a large teacher model into a smaller and efficient student model for model compression~\citep{gou2021knowledge}. It has been widely applied in computer vision~\citep{ahn2019variational, tiancontrastive2020,bergmann2020uninformed,zhao2022decoupled}, natural language processing~\citep{sanh2019distilbert,jiao2020tinybert, wang2020minilm, xu2024survey}, and speech recognition~\citep{chebotar2016distilling, fukuda2017efficient, tan2021towards} domains. 
In recent years, knowledge distillation has been a standard practice to enhance large language models (LLMs) with knowledge from more advanced LLMs, such as GPT-4~\citep{openai2023gpt}. This technique is typically used during the post-training stage of LLMs, where the student model learns directly using language modeling (LM) loss from a set of queries and responses generated by teacher LLMs. Post-training KD is simple and widely applicable, leading to the development of various advanced LLMs~\citep{alpaca,vicuna,sun2024conifer,cui2024ultrafeedback}, which significantly advances the development of LLMs. The success of post-training distillation raises the question of whether distillation LLMs in the pre-training stage is feasible.

\begin{figure}[t]
    \centering
\includegraphics[width=0.7\linewidth]{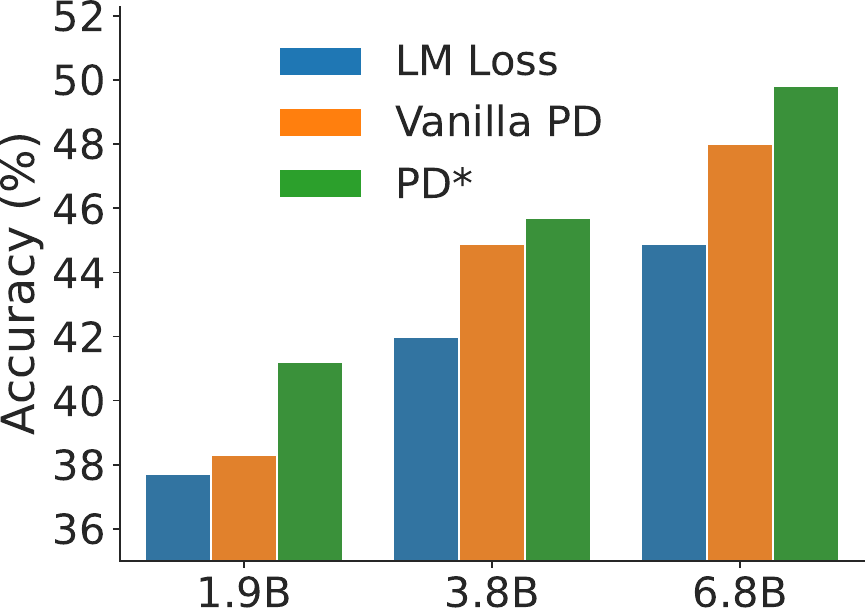}
    \caption{Results of the pre-trained 1.9B, 3.8B, and 6.8B student LLMs, using only LM loss, vanilla PD configuration (\cref{sec:preliminary_exp}), and a better PD configuration (PD$^*$) after our exploration. Details are placed in \cref{sec:app_better}.}
    \label{fig:figure1}
\end{figure}

In this paper, we extend knowledge distillation to the pre-training phase of LLMs, named pre-training distillation (PD). We primarily investigate pre-training with \textbf{logits-based} KD~\citep{gou2021knowledge}, where the student model learns from the teacher model generated logits of each token in the pre-training corpora using a KD loss, such as Kullback–Leibler divergence. The intuition is that the logits from the teacher model contain richer information and can serve as label smoothing~\citep{gou2021knowledge}, which could potentially accelerate the training of the student LLM and enhance its performance. Although the potential advantage of pre-training distillation is clear, 
there is limited exploration on how to better apply PD. Therefore, in this paper, we take an initial step in exploring the design space of pre-training distillation. Considering the key factors impacting distillation, we explore the design space of PD in four aspects: 
(1) \textbf{Logits processing}, focusing on the post-processing of the teacher LLM's logits to reduce the memory overhead, including truncation and normalization. (2) \textbf{Loss selection}, focusing on the choice of pre-training distillation loss. (3) \textbf{Scaling law}, covering varying sizes of student and teacher LLMs, as well as pre-training corpus size. (4) \textbf{\textit{Offline} or \textit{online}}, meaning logits are generated either from a pre-trained teacher LLM (\textit{offline}) or simultaneously during the pre-training of teacher LLM (\textit{online}). Figure~\ref{fig:figure1} illustrates the effectiveness of the explored better PD configuration (PD$^*$).

We conduct extensive experiments to explore the design space of PD.
Specifically, we first conduct a preliminary study using GLM-4-9B~\citep{glm2024chatglm} as the teacher model to generate logits for $100$ billion tokens, distilling a 1.9B student LLM from scratch using negative log-likelihood loss. Due to the large vocabulary size (about $150$k) of GLM-4-9B, we truncate the logits using top-$p$-$k$ truncation to reduce storage space: first using top-$p$~\citep{holtzmancurious} truncation with $p=0.95$, followed by top-$100$ truncation. The truncation reduces storage space by $4,000 \times$ to about 15 TB of disk space. 
The preliminary PD yields an average performance improvement of $1.6\%$ across a comprehensive set of
English and Chinese datasets, compared to standard pre-training with LM loss, which demonstrates the effectiveness of PD.
Based on the preliminary experiment, we explore the design space of PD using controlled experiments: (1) \textbf{Logits processing}. We investigate the impact of different $p$ and $k$ values on top-$p$-$k$ truncation results, and different normalization temperatures. We find no significant difference between various $p$ and $k$ values, with smaller $p$ or $k$ effectively reducing logits storage. The temperature for normalization should not be too high, and adaptive temperature shows no significant benefit. 
(2) \textbf{Loss selection}. We explore the choice of KD loss and the combination of KD loss with LM loss. We find that Kullback–Leibler divergence and negative log-likelihood loss result in similar improvements, but MSE loss suffers a significant drop. The best combination of LM and KD loss is using the Warmup-Stable-Decay (WSD; \citealp{hu2024minicpm}) method to schedule for the proportion of KD loss, paired with a WSD learning rate scheduler. This suggests that using a higher proportion of KD loss when maintaining a maximum learning rate can enhance model performance.
(3) \textbf{Scaling law}. We find that larger student LLMs generally benefit more from pre-training distillation, and a larger teacher LLM does not necessarily guarantee better results, potentially due to the capacity gap between student and teacher LLMs~\citep{mirzadeh2020improved}. We further conduct PD using $500$ billion tokens, and find
the improvement of PD is generally consistent. 
(4) \textbf{\textit{Offline} or \textit{online}}.
We observe that using \textit{online} logits for PD also yields improvement, 
although not as significant as \textit{offline} logits. 
% We suggest that one can adopt \textit{online} logits with no additional inference cost when pre-training a series of LLMs.
This suggests that one can save \textit{online} logits on the fly during pre-training with no additional inference cost for PD on a series of smaller LLMs.
In summary, we hope that our thorough exploration of the pre-training distillation design space will contribute to future practices.

\section{Design Space for PD}

% notations & formulation

% 介绍影响的factor: (1) logits (2) loss (3) scaling law
Considering a text $\vx = \{x_t\}_{t=1}^{T}$, a student LLM parameterized by $\theta_S$, and a teacher LLM parameterized by $\theta_T$,
we formalize the objective of distillation pretraining as follows:
\begin{equation}
\label{eq:eq1}
\small
    \theta_S^* = \text{arg min}_{\theta_S} \mathcal{L} = \text{arg min}_{\theta_S} [(1-\alpha)\mathcal{L}_{\text{lm}} + \alpha \mathcal{L}_{\text{kd}}]
\end{equation}
$\mathcal{L}_{\text{lm}}$ denotes the traditional one-hot language modeling pretraining loss, which can be formalized as:
\begin{equation}
\label{eq:eq2}
\small
\mathcal{L}_{\text{lm}} = \frac{1}{T}\sum_{t=1}^{T} -\log P_{\theta_S}(x_t|\vx_{<t})
\end{equation}
$\mathcal{L}_{\text{kd}}$ denotes the distillation loss, which can be formalized as:
\begin{equation}
\label{eq:eq3}
\small
\mathcal{L}_{\text{kd}} = \frac{1}{T}\sum_{t=1}^{T} L(P_{\theta_S}(x_t|\vx_{<t}), F(P_{\theta_T}(x_t|\vx_{<t})))
\end{equation}
$L$ denotes the distillation loss function, such as Kullback–Leibler divergence. $P_{\theta_S}$ and $P_{\theta_T}$ represent probability of the student and the teacher LLM, respectively. $F$ represents truncation and normalization operations conducted on the teacher LLM's logits, and $\tau$ is the temperature for normalization.
\begin{equation}
\label{eq:eq4}
\small
F(\vz) = \text{softmax}(\frac{\text{Truncate}(\vz)}{\tau})
\end{equation}

% \begin{figure}
%     \centering
%     \includegraphics[width=0.8\linewidth]{figs/design_space.pdf}
%     \caption{Design space for pre-training distillation.}
%     \label{fig:design_space}
% \end{figure}

Considering the key factors in Equation~\ref{eq:eq1}, we explore the design space of pre-training distillation in four dimensions: (1) The method $F$ for processing the teacher LLM logits, including the truncation method and temperature $\tau$ for normalization. (2) The choice of loss function, including the selection of distillation loss function
$L$ and the combination factor $\alpha$ of language modeling loss and distillation loss. (3) The scaling law of pre-training distillation, including the size of student and teacher LLMs, as well as the corpus size for pre-training the student LLM.
(4) The strategy of obtaining $P_{\theta_T}(x_t|\vx_{<t})$, either \textit{offline}, i.e., the logits generated from the pre-trained teacher LLM, or \textit{online}, i.e., the logits generated simultaneously during the teacher LLM's pre-training.
In this work, we aim to conduct a systematic empirical study to investigate the impact of these four aspects on pre-training distillation and inform future practices in pre-training distillation. 

% Furthermore, we aim to investigate the scaling law of distillation pretraining to explore the impact of varying sizes of student and teacher LLMs.

\section{Experiments}
In this section, we conduct a preliminary experiment to introduce the basic experimental settings of pre-training distillation and validate the efficacy of pre-training distillation (\cref{sec:preliminary_exp}) and empirical studies for these four main design dimensions of pre-training distillation for LLMs (\Cref{sec:loss_exp,sec:logits_exp,sec:online_exp,sec:scaling_law}).

\subsection{Preliminary Experiment}
\label{sec:preliminary_exp}
We first conduct a preliminary experiment to validate the feasibility of pre-training distillation. We use GLM-4-9B as the teacher LLM to distill of a 1.9B student LLM from scratch. To enhance training efficiency, we employ a two-stage paradigm: (1) store the teacher LLM's generated logits on the disk, (2) use these logits to train the student LLM.

\paragraph{Experimental Setup}
We first pre-train a 1.9B student LLM using pre-training distillation, namely \textbf{LLM-KD}. Specifically, we randomly sample $100$ billion tokens as pre-training data. We then obtain their logits from the teacher LLM and keep the text chunk size as $4096$, which is the same as the pre-training context length of the student LLM.
Due to the large vocabulary size (approximately $150$k items),
storing the logits of the whole vocabulary using float32 requires around $58.6$ PB of disk space, which is unaffordable. To reduce storage resources, we truncate the logits: we first select the top-$p$~\citep{holtzmancurious} logits with $p=0.95$, and then use top-$k$ truncation with $k=100$, resulting in a $4,000 \times$ reduced storage requirement of approximately 15 TB disk space for the $100$B tokens. We re-normalize the logits with temperature $\tau = 1.0$.
We use negative log-likelihood loss to conduct pre-training distillation, i.e., set $\alpha = 1$ in Equation~\ref{eq:eq1} and set $L = -F(P_{\theta_T}(x_t|\vx_{<t}))\log P_{\theta_S}(x_t|\vx_{<t})$ in Equation~\ref{eq:eq3}, where $F$ denotes our logits truncation method with a re-normalization with temperature $\tau = 1.0$.
Given the limited capacity of the student LLM, its performance on some evaluation datasets, such as MMLU~\citep{hendrycksmeasuring} and C-Eval~\citep{huang2024c}, is close to random guessing, making the results incomparable. Therefore, we conduct supervised fine-tuning (SFT; \citealp{ouyang2022training})
with additional $10$B high-quality instruct-tuning data after pre-training. 
In the SFT stage of these 10B tokens, we employ only language modeling loss rather, i.e., set $\alpha = 0$ in Equation~\ref{eq:eq1}. We employ the same settings as in pre-training distillation, except that we only use language modeling (LM) loss for pre-training a baseline 1.9B LLM for comparison, namely \textbf{LLM-LM}.
We conduct pre-training with Adam optimizer~\citep{kingma2014adam}, $2,048$ batch size, $4,096$ max sequence length, a cosine learning rate scheduler with $6\times10^{-4}$ maximum learning rate, $6\times10^{-5}$ minimum learning rate, and $1\%$ warmup rate. More experimental details are placed in \cref{sec:app_preliminary}.
% This is because the teacher LLM has not been trained on these instruct-tuning data, potentially resulting in a distribution discrepancy that renders its logits sub-optimal~\citep{turuvekere2024llm}. 

\paragraph{Evaluation Datasets}
% few-shot？
% 
We select several representative datasets to evaluate the pre-trained LLMs, including English language understanding and commonsense reasoning datasets: HellaSwag~\citep{zellers2019hellaswag}, WinoGrande~\citep{sakaguchi2020winogrande}, PIQA~\citep{bisk2020piqa}, MMLU~\citep{hendrycksmeasuring}; Chinese language understanding and commonsense reasoning datasets: KBQA~\citep{10.1007/978-3-319-50496-4_89, 10.1007/978-3-319-73618-1_86}, C3~\citep{sun2020investigating}, C-Eval~\citep{huang2024c}; and math dataset: GSM8k~\citep{cobbe2021training}.
When conducting evaluation, the sampling temperature is set to $0$. More evaluation details are shown in \cref{sec:app_preliminary}.

\paragraph{Experimental Results}
\begin{table*}[t]
    \centering
    \small
    \begin{adjustbox}{max width=1\linewidth}{
    \begin{tabular}{crrrrrrrrr}
    \toprule
    & HellaSwag & WinoGrande & PIQA & MMLU & KBQA & C3 & C-Eval & GSM8k & Average \\
    \midrule
    LLM-LM &$53.3$&$54.8$&$72.9$&$28.0$&$3.6$&$54.7$&$25.9$&$8.6$&$37.7$\\
    LLM-KD &$54.2$&$55.2$&$72.5$&$27.8$&$3.5$&$55.8$&$26.7$&$10.8$&$38.3$\\
$\Delta$&$\textcolor{darkgreen}{\uparrow {1.7}\%}$&$\textcolor{darkgreen}{\uparrow {0.7}\%}$&$\textcolor{red}{\downarrow {0.5}\%}$&$\textcolor{red}{\downarrow {0.5}\%}$&$\textcolor{red}{\downarrow {1.3}\%}$&$\textcolor{darkgreen}{\uparrow {1.9}\%}$&$\textcolor{darkgreen}{\uparrow {3.2}\%}$&$\textcolor{darkgreen}{\uparrow {24.6}\%}$&$\textcolor{darkgreen}{\uparrow {1.6}\%}$\\
    \bottomrule
    \end{tabular}
    }
\end{adjustbox}
    \caption{Preliminary experimental results on the evaluation datasets. $\Delta$ is relative to LLM-LM.}
    \label{tab:preliminary_result}
\end{table*}

% \begin{figure}
%     \centering
%     \includegraphics[width=0.6\linewidth]{figs/preliminary_loss.png}
%     \caption{Language modeling loss curves during the pre-training of LLM-LM and LLM-KD. We use distillation loss to pre-train LLM-KD and also record its language modeling loss during pre-training.}
%     \label{fig:preliminary_loss}
% \end{figure}

% 差一个对后文loss解释的引用
% 差一个对前文design space的引用

% The language modeling loss curves during the pre-training of LLM-LM and LLM-KD are shown in Figure~\ref{fig:preliminary_loss}. We can observe that LLM-KD's LM loss lags behind LLM-LM in the later stage of pre-training. At the end of pre-training, the loss for LLM-LM is $2.43$, while the loss for LLM-KD is $2.54$. 
The performance of pre-trained LLM-LM and LLM-KD is presented in Table~\ref{tab:preliminary_result}. 
We can observe that generally LLM-KD performs better than LLM-LM, though the improvement is marginal, indicating that pre-training distillation is feasible, but the current distillation configurations may not be optimal. Therefore, in the following sections (\Cref{sec:loss_exp,sec:logits_exp,sec:online_exp,sec:scaling_law}), we will explore the design space of pre-training distillation to identify more effective configurations.

\subsection{Design Dimension \#1: Logits Processing}
\label{sec:logits_exp}
This section explores the impact of logit processing in pre-training distillation, specifically $F$ in Equation~\ref{eq:eq1}, including the method for truncating logits and the temperature $\tau$ for normalization. If not stated otherwise, all experiments adopt the same setup as the preliminary experiment, except for the processing of logits. More experimental details and results are placed in \cref{sec:app_logits}.

\paragraph{Logits Truncation}
% \begin{table}[]
%     \centering
%     \small
%     \begin{adjustbox}{max width=1\linewidth}{
%     \begin{tabular}{lrrrrrr}
%     \toprule
%     $p$ & $0.95$ & $0.9$ & $0.8$ & $0.7$ & $0.6$ & $0.5$\\
%     \midrule
%     vanilla top-$p$ & $266$ & $125$ & $46$ & $21$ & $11$ & $6$ \\
%      top-$p$-$100$ & $37$ & $27$ & $16$ & $10$ & $7$ & $4$ \\
    
%     \bottomrule
%     \end{tabular}
%         }
% \end{adjustbox}
%     \caption{Logits sizes per token with difference $p$. The sizes are estimated using $10$ million tokens.}
%     \label{tab:logit_size}
% \end{table}

% \begin{figure}
%     \centering
% \includegraphics[width=0.4\linewidth]{figs/toppk.pdf}
%     \caption{Relative improvements compared to LLM-LM using different $p$ and $k$ in top-$p$-$k$ logits truncation.}
%     \label{fig:toppk}
% \end{figure}

\begin{figure}
    \centering
\includegraphics[width=0.8\linewidth]{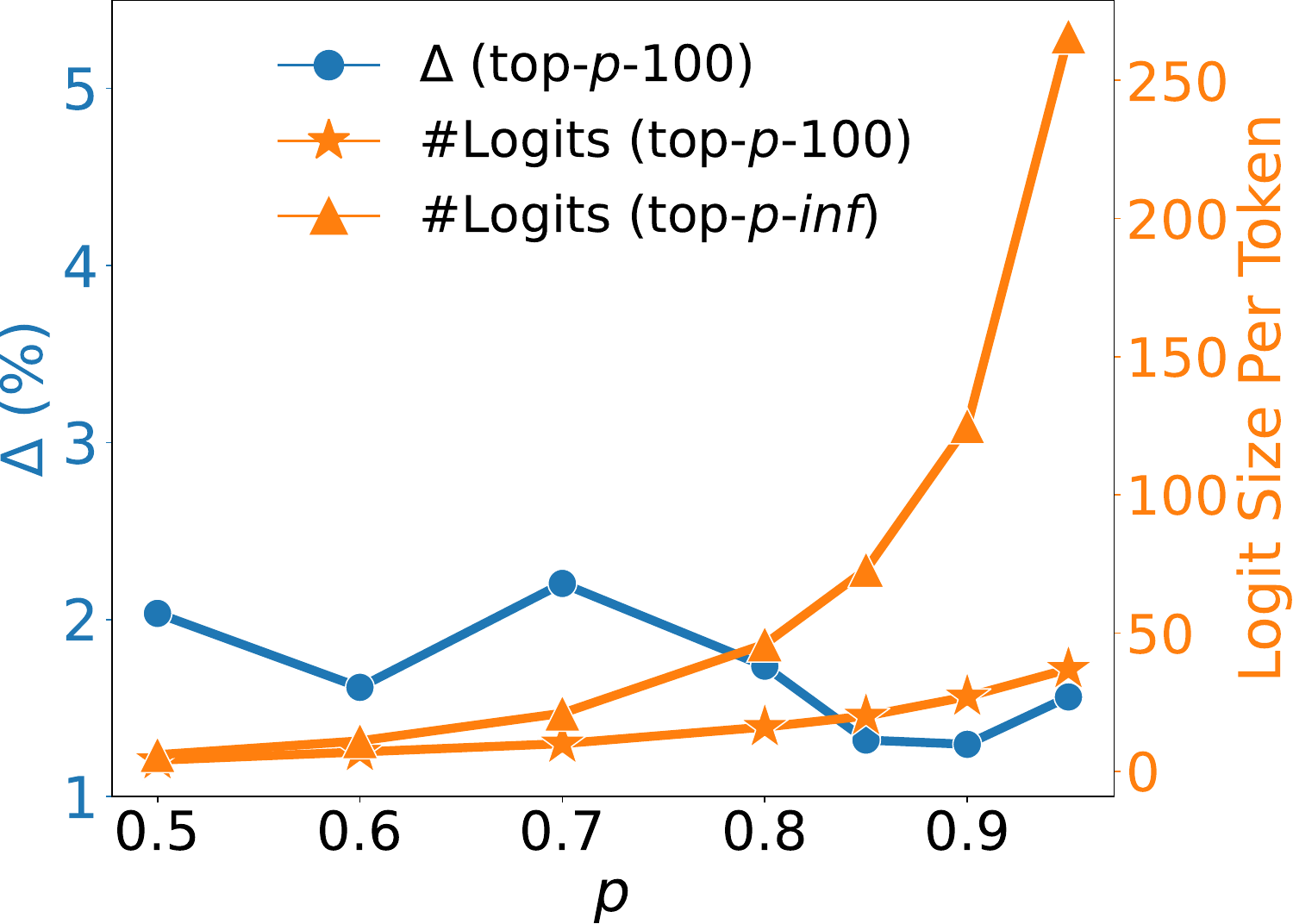}
    \caption{Relative improvements compared to LLM-LM using different $p$ in top-$p$-$100$ logits truncation and logits sizes per token with different $p$. The sizes are estimated using $10$ million tokens.}
    \label{fig:topp}
\end{figure}

\begin{figure}
    \centering
\includegraphics[width=0.8\linewidth]{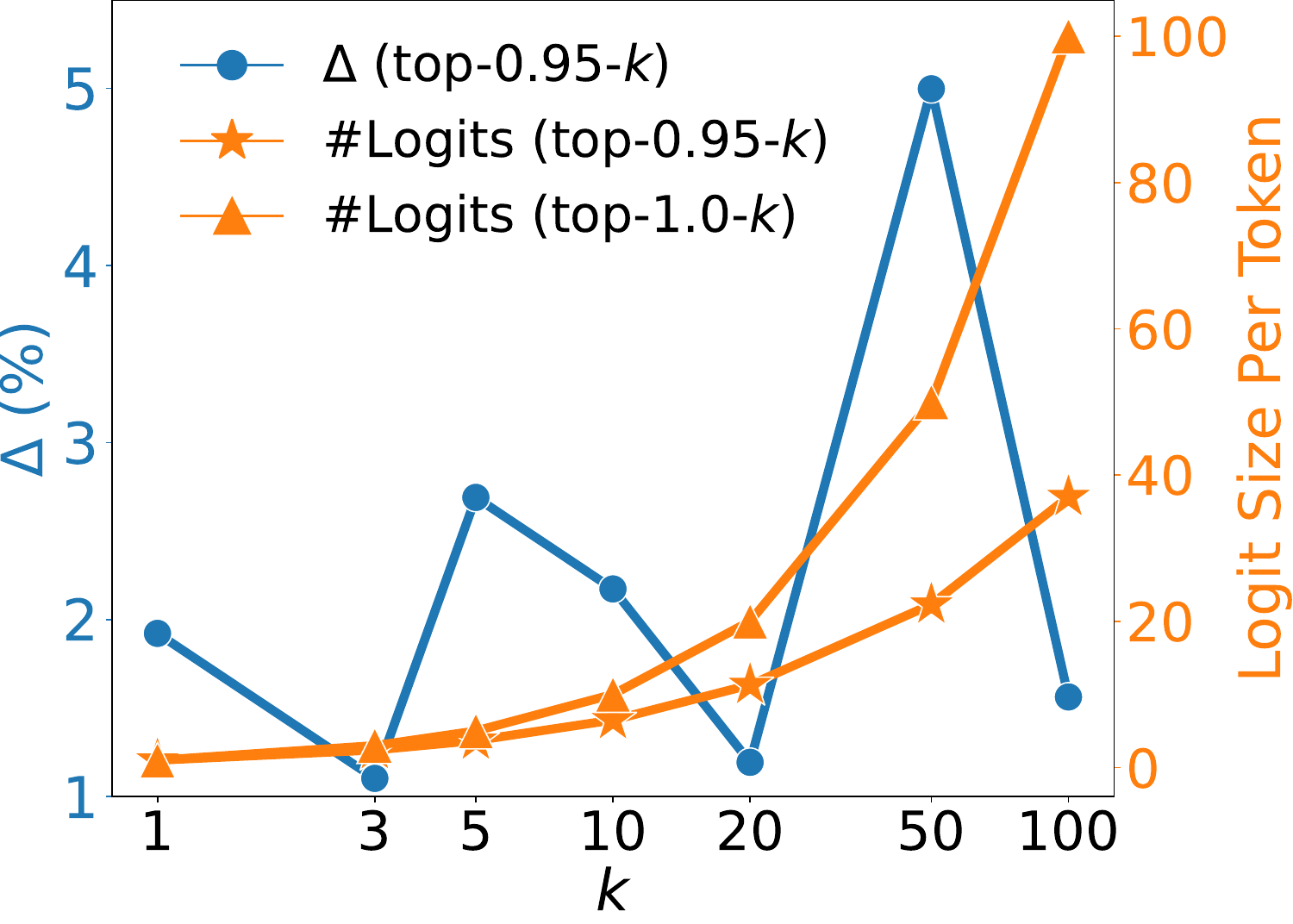}
    \caption{Relative improvements compared to LLM-LM using different $k$ in top-$0.95$-$k$ logits truncation and logits sizes per token with different $k$.}
    \label{fig:topk}
\end{figure}

As mentioned in the preliminary experiment (\cref{sec:preliminary_exp}), storing the logits of the entire vocabulary requires significant disk storage space. To save resources, we design a two-stage top-$p$-$k$ truncation method: truncating with top-$p$ first, followed by top-$k$ truncation. When the logits distribution is sharp, top-$p$ truncation is enough; when the distribution is more uniform with long-tailed non-trivial values, top-$k$ truncation works as a secondary truncation. 
Compared to vanilla top-$p$ and top-$k$ truncation, the top-$p$-$k$ method significantly reduces storage space, as shown in \Cref{fig:topp,fig:topk}. In this section, we empirically investigate the impact of different $p$ and $k$. Specifically, we set $k=100$ to study the impact of varying $p$ on top-$p$-$100$ truncation, and set $p=0.95$ to analyze the effect of different $k$ values on top-$0.95$-$k$ truncation. The results are shown in \Cref{fig:topp,fig:topk}. We can observe that (1) for top-$p$-$100$ truncation, different $p$ leads to similar improvements. A possible explanation is that in distillation pre-training, student LLM primarily captures the mass of the logits. This suggests that a smaller $p$ can be used to further reduce storage space.
(2) For top-$0.95$-$k$ truncation, all values of $k$ lead to improvements, with $k=50$ yielding the best results. For $k=1$, which is adopted in AFM pre-training~\citep{gunter2024apple}, is equivalent to using the LM loss but with labels generated from the teacher LLM and also yields an improvement. This may be due to the teacher LLM conducting implicit noise filtering in pre-training corpora. In general, pre-training distillation with different $p$ and $k$ values in top-$p$-$k$ truncation shows improvements with limited differences, and one can adopt smaller $p$ and $k$ in logits truncation to save storage disk space.

\paragraph{Temperature $\tau$}
\begin{table}[t]
    \centering
    \small
    \begin{adjustbox}{max width=1\linewidth}{
    \begin{tabular}{lrrrrrrrr}
    \toprule
    $\tau$ & $0.05$ & $0.1$ & $0.2$ & $0.5$ & $1.0$ & $2.0$ & $5.0$ & $10.0$\\
    \midrule
    ${\uparrow}$ & $1.6$ & $2.1$ & $2.5$ & $2.7$ & $1.6$ & $2.5$ & $-0.1$ & $1.0$\\
    \bottomrule
    \end{tabular}
        }
\end{adjustbox}
    \caption{Relative improvements (\%) compared to LLM-LM using different $\tau$ in logits normalization.}
    \label{tab:temperature}
\end{table}
\begin{table*}[t]
    \centering
    \small
    \begin{adjustbox}{max width=1\linewidth}{
    \begin{tabular}{lrrrrrrrrrr}
    \toprule
    & HellaSwag & WinoGrande & PIQA & MMLU & KBQA & C3 & C-Eval & GSM8k & Average & $\Delta$ \\
    \midrule
    NormKD &$51.2$&$54.1$&$71.0$&$26.6$&$3.2$&$54.6$&$29.0$&$8.0$&$37.2$ &$\textcolor{red}{\downarrow {1.3}\%}$\\
    WTTM &$51.4$&$56.2$&$72.9$&$26.7$&$3.6$&$55.1$&$27.3$&$9.2$&$37.8$&$\textcolor{darkgreen}{\uparrow {0.2}\%}$\\
    AdaKD\textsubscript{SD} &$54.7$&$54.5$&$73.0$&$25.7$&$3.7$&$56.1$&$25.9$&$11.8$&$38.2$&$\textcolor{darkgreen}{\uparrow {1.2}\%}$\\
    AdaKD\textsubscript{H} &$54.7$&$57.7$&$73.4$&$25.6$&$3.7$&$57.0$&$27.0$&$10.9$&$38.8$&$\textcolor{darkgreen}{\uparrow {2.8}\%}$\\
    \bottomrule
    \end{tabular}
    }
\end{adjustbox}
    \caption{Experimental results of LLMs pre-trained with different adaptive temperature $\tau$ methods.}
    \label{tab:ada_t}
\end{table*}

Another factor is the temperature $\tau$ in logits normalization. A lower temperature sharpens the logits distribution, while a higher temperature results in a more uniform distribution. We first examine the impact of different static $\tau$, as shown in Table~\ref{tab:temperature}. We can observe that lower temperatures ($\tau \leq 2.0$) lead to similar improvement, whereas at higher temperatures ($\tau \geq 5.0$), the improvement is limited. This suggests that learning from a more uniform distribution may be not efficient for student LLM. 
We also explore adaptive temperature, where temperature dynamically adjusts based on each sample, i.e., each token in pre-training distillation. We investigate two representative methods: NormKD~\citep{chi2023normkd} and WTTM~\citep{zhengknowledge}. NormKD applies adaptive temperature to both teacher and student logits, while WTTM applies only to the teacher logits. 
In this experiment, along with temperature $\tau$, the loss calculation method is also modified. For details, refer to their original papers, and relevant hyper-parameters in loss calculation are listed in \cref{sec:app_logits}. 
We also implement a compact version of the adaptive temperature method, named AdaKD, which applies a higher temperature to smooth sharper teacher logits and a lower temperature for less sharp logits
to help the student LLM focus on the most important parts~\citep{wei2024dynamic}. We use standard deviation and entropy to measure the sharpness of the logits, referred to as AdaKD\textsubscript{SD} and AdaKD\textsubscript{H}, and adaptively calculate the temperature accordingly. 
AdaKD\textsubscript{SD} adopts the standard deviation as the temperature $\tau$. AdaKD\textsubscript{H} adopts $\tau_H$ in Equation~\ref{eq:entropy}.
\begin{equation}
\small
\label{eq:entropy}
\tau_H = \tau_{\text{max}} - (\tau_{\text{max}} - \tau_{\text{min}}) \times \frac{H}{H_{\text{max}}}
\end{equation}
$H$ denotes the entropy of each sample. More experimental details are placed in \cref{sec:app_logits}. The results are presented in Table~\ref{tab:ada_t}. We can observe that AdaKD\textsubscript{H} performs the best, but compared to static temperature ($\tau=0.5$), adaptive temperature does not show significant additional improvement.

\subsection{Design Dimension \#2: Loss Selection}
\label{sec:loss_exp}

% \begin{figure}
%     \centering
%     \includegraphics[width=0.9\linewidth]{figs/loss.pdf}
%     \caption{Caption}
%     \label{fig:enter-label}
% \end{figure}

This section explores loss selection in pre-training distillation, including the types of distillation loss $L$ and the selection of combinations with the LM loss, i.e., $\alpha$ in Equation~\ref{eq:eq1}. For all the experiments in this section, all settings remain the same as those in the preliminary experiment, except for the choice of loss. More results are placed in \cref{sec:app_loss}.

\paragraph{Distillation Loss Function $L$}

We first explore the impacts of different distillation loss functions $L$. Specifically, we examine three common-used types of loss function: negative log-likelihood (NLL) as used in the preliminary experiment (\cref{sec:preliminary_exp}), Kullback–Leibler divergence (KLD), and mean squared error (MSE) loss. To control for variables, we omit LM loss and only use the distillation loss, setting $\alpha = 1$ in Equation~\ref{eq:eq1}.
The experimental results are presented in Table~\ref{tab:loss_selection}. We can find that the LLMs trained with NLL and KLD loss both perform better than LLM-LM. While LLM-KLD generally outperforms LLM-NLL, the latter demonstrates superior performance on more challenging datasets, such as MMLU and C-Eval. The student LLM trained with MSE loss exhibits a significant performance decline, as observed in previous studies~\citep{muralidharan2024compact}. This finding contrasts with prior research in image classification~\citep{kimcomparing}, which finds MSE loss is the most superior choice in knowledge distillation, indicating that the pre-training distillation of LLMs involves new training dynamics and requires further investigation.
% This may be due to MSE loss potentially overlooking the next token's subtle but important distributional information. 

\begin{table}[t]
    \centering
    \small
    \begin{adjustbox}{max width=1\linewidth}{
    \begin{tabular}{lrrrrrrrr}
    \toprule
    $\alpha$ & $0.1$ & $0.5$ & $0.6$ & $0.7$ & $0.8$ & $0.9$ & $0.95$ & $1.0$\\
    \midrule
    ${\uparrow}$ & $0.1$ & $1.5$ & $1.4$ & $2.9$ & $2.0$ & $3.6$ & $2.5$ & $1.6$\\
    \bottomrule
    \end{tabular}
        }
\end{adjustbox}
    \caption{Relative improvements (\%) compared to LLM-LM using different $\alpha$ in combination of $\mathcal{L_{\text{lm}}}$ and $\mathcal{L_{\text{kd}}}$.}
    \label{tab:loss_alpha}
\end{table}

\begin{table*}[t]
    \centering
    \small
    \begin{adjustbox}{max width=1\linewidth}{
    \begin{tabular}{lrrrrrrrrrr}
    \toprule
    & HellaSwag & WinoGrande & PIQA & MMLU & KBQA & C3 & C-Eval & GSM8k & Average & $\Delta$ \\
    \midrule
    $0$-$\alpha$+WSD-LR     &$54.1$&$55.1$&$73.1$&$27.5$&$3.8$&$55.6$&$27.5$&$8.5$&$38.2$&$\textcolor{darkgreen}{\uparrow {1.2}\%}$\\
    \midrule
    LLM-NLL&$54.2$&$55.2$&$72.5$&$27.8$&$3.5$&$55.8$&$26.7$&$10.8$&$38.3$&$\textcolor{darkgreen}{\uparrow {1.6}\%}$\\
    LLM-KLD &$55.3$&$56.7$&$73.5$&$26.7$&$3.6$&$56.7$&$25.4$&$11.5$&$38.7$ & $\textcolor{darkgreen}{\uparrow {2.6}\%}$\\
    LLM-MSE &$44.6$&$55.0$&$69.6$&$25.2$&$2.8$&$52.2$&$25.6$&$3.9$&$34.9$&$\textcolor{red}{\downarrow {7.6}\%}$\\
    \midrule
    Linear Inc &$53.6$&$55.2$&$73.1$&$25.9$&$3.4$&$56.4$&$28.9$&$8.5$&$38.1$&$\textcolor{darkgreen}{\uparrow {1.1}\%}$\\
    Linear Dec &$53.4$&$56.6$&$72.9$&$29.6$&$3.6$&$56.0$&$30.5$&$11.4$&$39.2$&$\textcolor{darkgreen}{\uparrow {4.1}\%}$\\
    Period &$52.9$&$55.0$&$72.3$&$28.4$&$3.4$&$55.1$&$27.9$&$9.4$&$38.0$&$\textcolor{darkgreen}{\uparrow {0.9}\%}$\\
    $1$-$\alpha$+WSD-LR
&$56.1$&$57.2$&$73.6$&$27.0$&$3.8$&$58.3$&$29.1$&$11.6$&$39.6$&$\textcolor{darkgreen}{\uparrow {5.0}\%}$\\
    WSD-$\alpha$+Cos-LR &$54.0$&$55.4$&$72.7$&$25.1$&$3.7$&$57.6$&$29.4$&$10.6$&$38.6$&$\textcolor{darkgreen}{\uparrow {2.3}\%}$\\
    WSD-$\beta$+WSD-LR &$53.1$&$55.2$&$73.7$&$27.5$&$3.6$&$55.7$&$25.0$&$11.2$&$38.1$&$\textcolor{darkgreen}{\uparrow {1.1}\%}$\\
    WSD-$\alpha$+WSD-LR &$56.4$&$57.7$&$73.6$&$31.8$&$2.6$&$57.6$&$33.8$&$12.5$&$40.7$&$\textcolor{darkgreen}{\uparrow {8.0}\%}$\\
    \bottomrule
    \end{tabular}
    }
\end{adjustbox}
    \caption{Experimental results of LLMs pre-trained with different pre-training loss. $\Delta$ is relative to LLM-LM. $0$-$\alpha$ and $1$-$\alpha$ denote setting $\alpha = 0$ and $\alpha = 1.0$, respectively. $0$-$\alpha$+WSD-LR represents LLM-LM training with the WSD scheduler, which serves as a baseline. Cos-LR means a cosine learning rate scheduler. $\beta \equiv
 1 - \alpha$, and WSD-$\beta$ denotes applying the WSD scheduler to the proportion of LM loss.}
    \label{tab:loss_selection}
\end{table*}

\paragraph{Combination of $\mathcal{L_{\text{lm}}}$ and $\mathcal{L_{\text{kd}}}$}

We examine the impact of different combinations of $\mathcal{L_{\text{lm}}}$ and $\mathcal{L_{\text{kd}}}$. We set $\mathcal{L_{\text{kd}}}$ as negative log-likelihood loss for all experiments. Specifically, we first explore the effect of different values of static $\alpha$, ranging in \{$0.0$, $0.5$, $0.6$, $0.7$, $0.8$, $0.9$, $0.95$, $1.0$\}. The results are shown in Table~\ref{tab:loss_alpha}, and we can observe that as $\alpha$ increases, the PD performance improves generally, then declines, with the best performance at $\alpha = 0.9$. This suggests that while a higher proportion of distillation loss can boost the distillation performance, an appropriate ratio (about $10\%$) of LM loss can further enhance pre-training distillation performance. 

\begin{figure}
    \centering
    \includegraphics[width=0.7\linewidth]{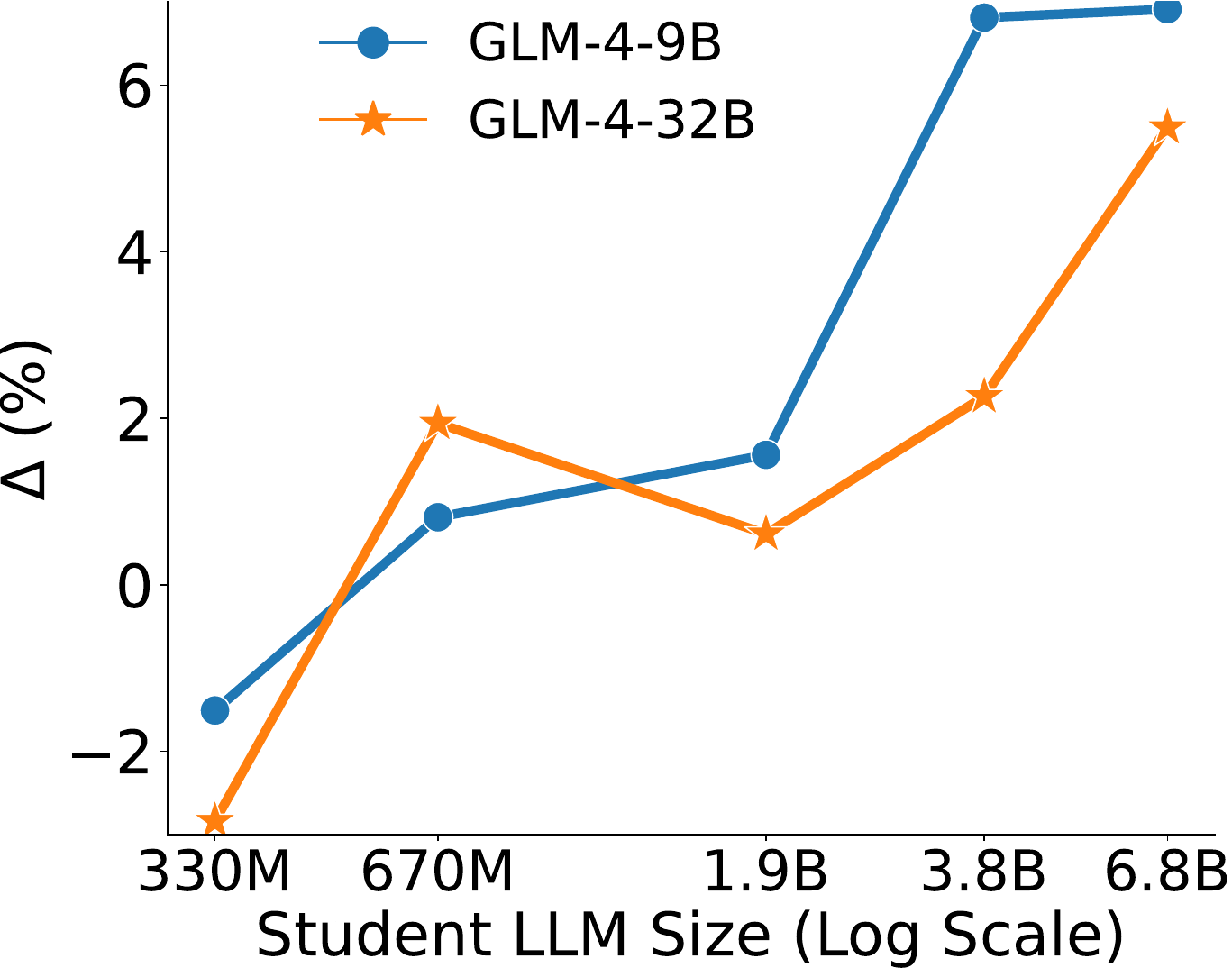}
    \caption{Relative improvements compared to LLM-LM using varying sizes of student and teacher LLMs.}
    \label{fig:model_size}
\end{figure}

We further explore dynamic scheduling of $\alpha$ in the following ways: (1) $\alpha$ linearly increases from $0$ to $1$, namely Linear Inc, or decreases from $1$ to $0$, namely, Linear Dec, during pre-training. The intuition of the former is that training initially with LM loss may help mitigate the effects of the capacity gap with the teacher LLM; the latter is that using KD loss first may provide better optimization initialization~\citep{yim2017gift}. (2) $\alpha$ periodically varies between $0$ and $0.9$, namely Period, setting $\alpha$ to $0.9$ at every fourth batch and $0$ for the other batches~\citep{kiefel2024lokilm}. (3) We employ a nonlinear scheduler, warmup-stable-decay (WSD; \citealp{hu2024minicpm}), for scheduling $\alpha$, namely WSD-$\alpha$. Specifically, we first linearly increase $\alpha$ from $0$ to $1.0$ during the warm up stage, then stay $\alpha=1.0$, and finally apply cosine decay to reduce $\alpha$ from $1.0$ to $0$. We set the warmup ratio at $10\%$ and the decay ratio at $1\%$. Furthermore, we employ the WSD learning rate scheduler~\citep{hu2024minicpm}, namely WSD-LR, setting its warmup and decay ratios as those of WSD-$\alpha$. The intuition is that when the learning rate stays at its maximum, utilizing KD loss may enhance training efficiency. The results are shown in Table~\ref{tab:loss_selection}. We can observe that: (1) A linear decrease in $\alpha$ outperforms a linear increase, indicating that involving more KD loss in the early pre-training stage is more beneficial. (2) The WSD learning rate scheduler generally provides benefits, with greater gains when combined with KD loss. (3) The WSD $\alpha$ scheduler with the WSD learning rate scheduler yields the best performance, and the improvement of WSD $\beta$ ($\beta \equiv
 1 - \alpha$) scheduler with WSD-LR is limited, suggesting that using KD loss when maintaining a high learning rate effectively enhances model performance. Compared to WSD-LR with only KD loss, WSD-$\alpha$ performs better, indicating that a small proportion of LM loss can further enhance distillation performance.

% Linear Inc
% Linear Dec
% Period
% Con-a + Cos-lr
% WSD-a + Cos-lr
% Con-a + WSD-lr
% WSD-a + WSD-lr

% and the WSD scheduler, where the latter aims to utilize NLL loss when the learning rate is at its maximum to enhance learning efficiency.

\subsection{Design Dimension \#3: Scaling Law}
\label{sec:scaling_law}

\begin{table*}[t]
    \centering
    \small
    \begin{adjustbox}{max width=1\linewidth}{
    \begin{tabular}{lrrrrrrrrrr}
    \toprule
    & HellaSwag & WinoGrande & PIQA & MMLU & KBQA & C3 & C-Eval & GSM8k & Average & $\Delta$ \\
    \midrule
    LLM-Online-100B-L &$30.1$&$53.0$&$62.1$&$24.5$&$0.7$&$40.2$&$25.9$&$2.4$&$29.8$&$\textcolor{red}{\downarrow {20.9}\%}$\\
    LLM-Online-100B &$49.5$&$54.2$&$70.5$&$25.2$&$3.0$&$54.2$&$25.5$&$8.0$&$36.3$&$\textcolor{red}{\downarrow {3.9}\%}$\\
    LLM-Online-100B* &$52.9$&$55.4$&$72.3$&$26.6$&$3.6$&$57.0$&$25.4$&$10.0$&$37.9$&$\textcolor{darkgreen}{\uparrow {0.5}\%}$\\
    \bottomrule
    \end{tabular}
    }
\end{adjustbox}
    \caption{Experimental results of different LLMs pre-trained with \textit{online} logits. $\Delta$ is relative to LLM-LM.}
    \label{tab:offline}
\end{table*}

We investigate the scaling law of pre-training distillation, including the impact of varying sizes of student and teacher LLMs, as well as the pre-training corpus size. All experimental settings are the same as the preliminary experiment, except for the sizes of LLMs and pre-training corpus. More experimental details are placed in \cref{sec:app_model_size}.

\paragraph{Model Size}
We first investigate the performance with varying sizes of student and teacher LLMs in pre-training distillation.
Specifically, we adopt teacher LLMs of 9B and 32B to distill student LLMs of 330M, 670M, 1.9B, 3.8B, and 6.8B. 
For each size of the student LLM, we pre-train a baseline LLM using only the LM loss, i.e., setting $\alpha=0$ in Equation~\ref{eq:eq1}. The relative improvements compared to baseline LLMs are illustrated in Figure~\ref{fig:model_size}. We can observe that: (1) Larger student LLMs generally benefit more from pre-training distillation. (2) Distilling from a larger teacher LLM does not necessarily yield better performance. This may be due to the capacity gap between teacher and student LLMs~\citep{mirzadeh2020improved,gou2021knowledge}. Increasing the student LLM size or using a smaller teacher LLM can both reduce this gap and hence improve distillation performance. From a compression perspective, larger LLMs compress information more effectively and achieve better compression rates~\citep{deletanglanguage}, potentially making it harder for smaller LLMs to learn. Our experiments demonstrate that pre-training distillation is effective when the size of the student LLM reaches about $10\%$ or more of the teacher LLM size, and as the proportion increases, the benefits of pre-training distillation grow until reaches the turning point. Due to computational constraints, we do not explore the turning point of performance gain to the proportion, which we leave as future work.
Furthermore, scaling the student LLM to larger sizes may yield new interesting findings, such as the weak-to-strong generalization~\citep{burnsweak}: using a small teacher LLM to help train a large student LLM. Due to computational constraints, we leave these explorations as future work. 

\paragraph{Corpus Size}
\begin{figure}
    \centering
    \includegraphics[width=0.75\linewidth]{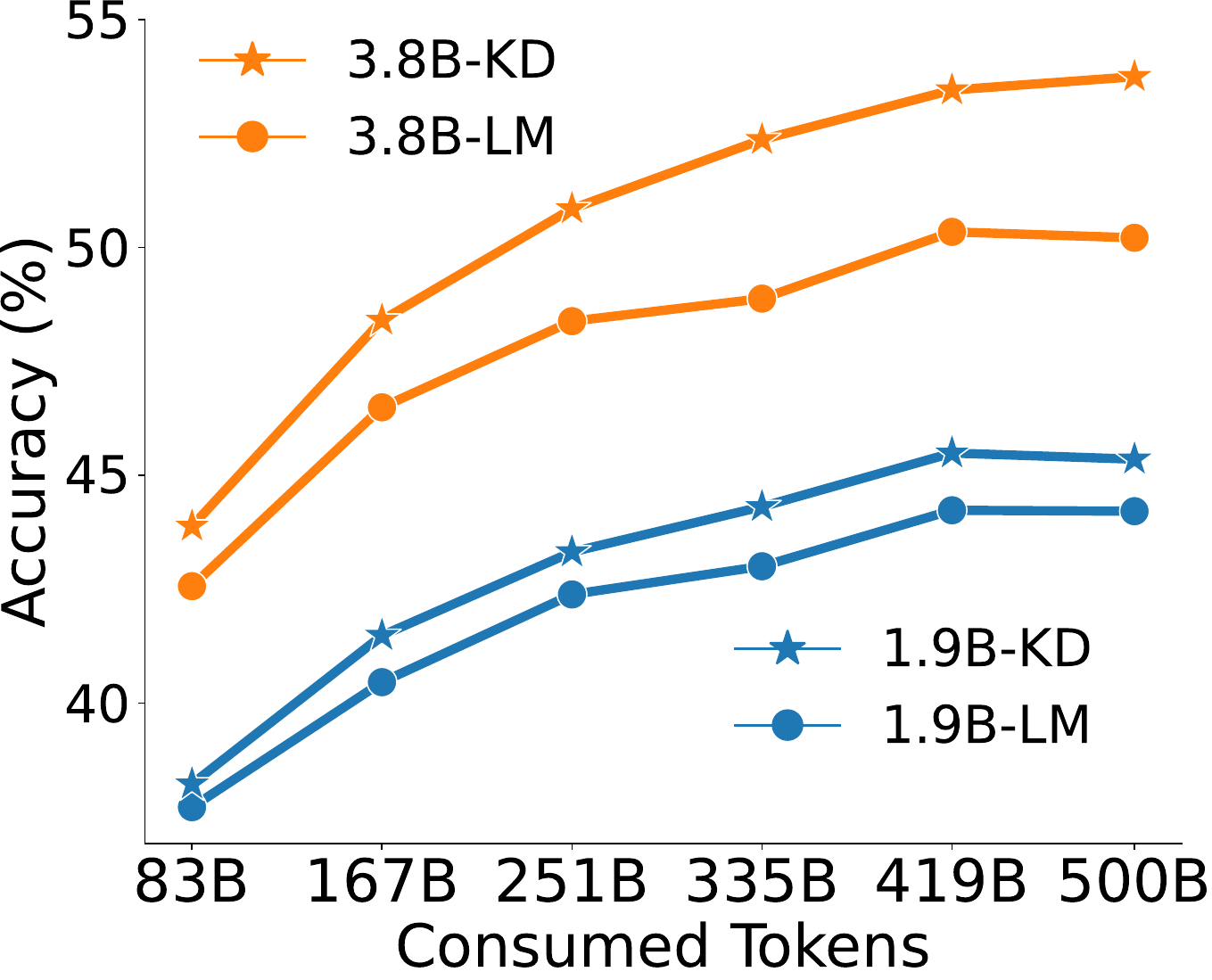}
    \caption{Experimental results of the checkpoints saved every $10,000$ step (about $83$B tokens) during the pre-training of 1.9B and 3.8B LLMs on 500B tokens. The last data point is from the checkpoint saved at the end.}
    \label{fig:corpus_size}
\end{figure}
We further investigate the impact of pre-training corpus size. Specifically, we use GLM-4-9B as the teacher LLM and distill 1.9B and 3.8B student LLMs with $500$ billion tokens. We also pre-training corresponding baseline LLMs with only LM loss. We save a checkpoint every $10,000$ optimization step (about $83$B tokens) and save the last checkpoint at the end of pre-training. All the other settings are consistent with the preliminary experiment. The results are illustrated in Figure~\ref{fig:corpus_size}.
We can observe that: (1) Compared to student LLMs trained only with LM loss, pre-training distillation consistently yields improvements throughout the pre-training process, remaining effective with more tokens. (2) 
The gains from pre-training distillation increase initially during pre-training and then converge with a slight decrease, and are still significant are the end of pre-training. This suggests that pre-training distillation not only enhances training efficiency but also improves the performance upper bound of student LLMs. Due to computational limitations, we do not reach trillion-level tokens for pre-training which are used by most advanced LLMs~\citep{team2024gemma,dubey2024llama,gunter2024apple,glm2024chatglm,qwen2.5,liu2024deepseek}. 
We believe that pre-training distillation is also effective using several trillion tokens and encourage future LLM development to incorporate pre-training distillation.

\subsection{Design Dimension \#4: \textit{Offline} or \textit{Online}}
\label{sec:online_exp}
% \footnote{This definition is different from ``On-Policy Distillation''~\citep{agarwal2024policy}, which trains the student on its self-generated output sequences.} 
% Besides the logits, the experimental setup is the same as the preliminary experiment (\cref{sec:preliminary_exp}).

This section explores how logits are obtained, either \textit{offline} or \textit{online}. \textit{Offline} means that logits are obtained from a pre-trained teacher LLM, which is the setting for all previous experiments. \textit{Online} refers to storing logits generated 
simultaneously during the pre-training of the teacher LLM. 
The advantage of \textit{online} is that 
it does not require additional inference from the teacher LLM if one stores the logits during teacher pre-training. 
Another potential advantage is that learning from \textit{online} logits is similar to curriculum learning~\citep{soviany2022curriculum}, which may help mitigate the capacity gap and improve learning efficiency. 
Due to the high cost of pre-training GLM-4-9B from scratch, we preliminarily pre-train GLM-4-9B from scratch using 400 billion tokens while storing the logits for each token. We 
first distill two 1.9B student LLMs using the setup in \cref{sec:preliminary_exp}: LLM-Online-100B-L and LLM-Online-100B, which adopt the first and the last 100 billion tokens during teacher LLM's pre-training process, respectively. Experimental details are presented in \cref{sec:app_online}. The results are presented in Table~\ref{tab:offline}. Both LLMs yield poor performance, particularly LLM-Online-100B-L. The reason may be that the teacher LLM is far from convergence, and hence the logits contain substantial noise. We adjust the loss calculation with $\alpha=0.1$ and use top-$0.95$-$50$ truncation to train LLM-Online-100B*, which performs slightly better than LLM-LM, although it still underperforms LLM-KD using \textit{offline} logits.
This indicates that even logits generated by a non-converged teacher LLM can help pre-training student LLM, suggesting that using \textit{online} logits is also effective and better practice is to utilize the logits from the later stages of the teacher LLM's pre-training.
We suggest that if one aims to pre-train only an LLM, using \textit{offline} logits of a pre-trained teacher LLM is better; if one
aims to pre-train a series of LLMs of varying sizes, one can first pre-train the largest LLM while storing \textit{online} logits, and then pre-train smaller LLMs with \textit{online} logits. 

% Therefore, we suggest that if one aims to pretrain a series of LLMs of varying sizes, one can first pretrain the largest LLM while storing its logits simultaneously, and then use those logits to pretrain smaller LLMs.

% best practices 1.5b
% dynamic temperature
% online distillation

% continual pre-trianing

% \input{04.discussion}
\section{Related Work}

Knowledge distillation aims to transfer knowledge from a large teacher model into a smaller student model for model compression. It is first formalized by \citet{hinton2015distilling}, which adopts the teacher model's logits as soft targets to train the student model, which can provide richer information~\citep{gou2021knowledge} and is also similar to label smoothing~\citep{kim2017transferring} and regulation~\citep{muller2019does, ding2019adaptive}. In this paper, we also focus on logits-based knowledge distillation. Knowledge distillation has been widely applied in in computer vision~\citep{komodakis2017paying, ahn2019variational, wang2020exclusivity,bergmann2020uninformed,zhao2022decoupled,habib2023knowledge}, natural language processing~\citep{sanh2019distilbert,jiao2020tinybert, wang2020minilm,chen2020distilling,alpaca,xu2024survey}, and speech recognition~\citep{chebotar2016distilling, fukuda2017efficient, tan2021towards} domains.

Since the emergence of ChatGPT~\citep{openai2022chatgpt}, knowledge distillation has become one of the most crucial techniques for enhancing large language models (LLMs). Typically, KD is applied during the post-training phase in sequence-level~\citep{kim2016sequence} to efficiently align them with humans~\citep{xu2024survey}, where student LLMs are trained using a teacher-forcing language modeling loss from instructions and corresponding responses generated by advanced proprietary LLMs, such as GPT-4~\citep{openai2023gpt}. Alpaca~\citep{alpaca} is the first public LLM distilled from ChatGPT, providing a practical approach for improving open-source LLMs. Due to the compactness and efficacy of post-training KD, it is widely applied in developing various LLMs~\citep{xu2023wizardlm, alpaca, vicuna, mitra2023orca, ding2023enhancing, sun2024conifer, qi2024adelie, cui2024ultrafeedback}, which significantly advances the development of LLMs.

For pre-training distillation of language models, there are two main categories of related work: (1) Distilling small language models in the pre-ChatGPT era~\citep{sanh2019distilbert,jiao2020tinybert,wang2020minilm,xu2020bert,sun2020mobilebert,zhang2020ternarybert,liu2020fastbert,hou2020dynabert}. These approaches are usually based on models with only several million parameters, such as BERT~\citep{kenton2019bert}, and hence their training configurations may not be directly applicable for billion-level LLMs. (2) Distilling LLMs~\citep{gu2024minillm,muralidharan2024compact,kiefel2024lokilm,turuvekere2024llm,team2024gemma,gunter2024apple}. MiniLLM~\citep{gu2024minillm} is trained based on a pre-trained LLM rather than from scratch. 
Gemma 2~\citep{team2024gemma}, AFM~\citep{gunter2024apple}, LokiLM~\citep{kiefel2024lokilm}, and Minitron~\citep{turuvekere2024llm} employ pre-training distillation but provide limited details on the distillation process. While \citet{muralidharan2024compact} explores the best practices for pruning and distillation of LLMs, it mainly focuses on pruning and does not systematically explore pre-training distillation. In this work, we systematically explore the design space of pre-training distillation and conduct extensive experiments to find key impact factors and better configurations. Our findings can also be applied to previous pruning and distillation work, and we hope these explorations could inform future practices in pre-training distillation.

% In recent years, knowledge distillation has been a standard practice to enhance large language models (LLMs) with knowledge from more advanced proprietary LLMs, such as GPT-4~\citep{openai2023gpt}. This technique is typically used during the post-training stage of LLMs in sequence-level~\citep{kim2016sequence} to efficiently align them with humans~\citep{xu2024survey}, where the student model learns directly using teacher-forcing language modeling loss (LM loss) from a set of queries and corresponding responses generated by advanced teacher LLMs. Sequence-level KD is simple and widely applicable, leading to the development of various open-source datasets and LLMs~\citep{xu2023wizardlm, alpaca, vicuna, mitra2023orca, ding2023enhancing, sun2024conifer, qi2024adelie, cui2024ultrafeedback}, which significantly advances the development of LLMs. The success of post-training distillation raises the question of whether distillation in the pre-training stage is feasible.
\section{Conclusion}

In this paper, we systematically explore the design space of pre-training distillation, including four main impacting factors: logits processing, loss selection, scaling law, and strategies for obtaining logits, i.e., \textit{offline} or \textit{online}. We conduct extensive experiments to study each design dimension and identify better configurations. We also draw some interesting conclusions, such as larger student LLMs generally benefiting more from pre-training distillation while larger teacher LLMs do not guarantee better results. We hope our exploration will inform future practices in pre-training distillation.

\section*{Limitations}

The main limitation of this work is that we do not explore the interactions between different factors in pre-training distillation, that is, the different combinations of factors. This is unaffordable, as these experiments are too resource-intensive given the complexity of factor combinations. Our controlled variable experiments have already incurred significant computational costs, which emit a significant amount of carbon dioxide and negatively impact the environment~\citep{strubell2019energy}. While searching the combinations of factors could identify best practices, we believe our experiments and explorations are sufficiently solid to inform future practices in pre-training distillation.

\section*{Ethical Considerations}

We discuss the ethical considerations of this work: (1) Intellectual property. We strictly adhere to the copyright licenses of all the used models and datasets. 
(2) Intended use. Our work explores the design space of pre-training distillation, aiming to inform future practices in pre-training distillation. 
(3) Potential risk control. We believe the data used has been properly anonymized. As an empirical study, we do not publish additional artifacts. 
(4) AI assistance. We adopt ChatGPT for paraphrasing some sentences and grammar checks.

\bibliography{custom}

\begin{thebibliography}{72}
\providecommand{\natexlab}[1]{#1}

\bibitem[{Ahn et~al.(2019)Ahn, Hu, Damianou, Lawrence, and Dai}]{ahn2019variational}
Sungsoo Ahn, Shell~Xu Hu, Andreas Damianou, Neil~D Lawrence, and Zhenwen Dai. 2019.
\newblock Variational information distillation for knowledge transfer.
\newblock In \emph{Proceedings of the IEEE/CVF conference on computer vision and pattern recognition}, pages 9163--9171.

\bibitem[{Ainslie et~al.(2023)Ainslie, Lee-Thorp, de~Jong, Zemlyanskiy, Lebron, and Sanghai}]{ainslie2023gqa}
Joshua Ainslie, James Lee-Thorp, Michiel de~Jong, Yury Zemlyanskiy, Federico Lebron, and Sumit Sanghai. 2023.
\newblock Gqa: Training generalized multi-query transformer models from multi-head checkpoints.
\newblock In \emph{Proceedings of EMNLP}, pages 4895--4901.

\bibitem[{Bergmann et~al.(2020)Bergmann, Fauser, Sattlegger, and Steger}]{bergmann2020uninformed}
Paul Bergmann, Michael Fauser, David Sattlegger, and Carsten Steger. 2020.
\newblock Uninformed students: Student-teacher anomaly detection with discriminative latent embeddings.
\newblock In \emph{Proceedings of the IEEE/CVF conference on computer vision and pattern recognition}, pages 4183--4192.

\bibitem[{Bisk et~al.(2020)Bisk, Zellers, Gao, Choi et~al.}]{bisk2020piqa}
Yonatan Bisk, Rowan Zellers, Jianfeng Gao, Yejin Choi, et~al. 2020.
\newblock Piqa: Reasoning about physical commonsense in natural language.
\newblock In \emph{Proceedings of AAAI}, volume~34, pages 7432--7439.

\bibitem[{Burns et~al.(2024)Burns, Izmailov, Kirchner, Baker, Gao, Aschenbrenner, Chen, Ecoffet, Joglekar, Leike et~al.}]{burnsweak}
Collin Burns, Pavel Izmailov, Jan~Hendrik Kirchner, Bowen Baker, Leo Gao, Leopold Aschenbrenner, Yining Chen, Adrien Ecoffet, Manas Joglekar, Jan Leike, et~al. 2024.
\newblock Weak-to-strong generalization: Eliciting strong capabilities with weak supervision.
\newblock In \emph{Proceedings of ICML}.

\bibitem[{Chebotar and Waters(2016)}]{chebotar2016distilling}
Yevgen Chebotar and Austin Waters. 2016.
\newblock Distilling knowledge from ensembles of neural networks for speech recognition.
\newblock In \emph{Interspeech}, pages 3439--3443.

\bibitem[{Chen et~al.(2020)Chen, Gan, Cheng, Liu, and Liu}]{chen2020distilling}
Yen-Chun Chen, Zhe Gan, Yu~Cheng, Jingzhou Liu, and Jingjing Liu. 2020.
\newblock Distilling knowledge learned in bert for text generation.
\newblock In \emph{Proceedings of the 58th Annual Meeting of the Association for Computational Linguistics}, pages 7893--7905.

\bibitem[{Chi et~al.(2023)Chi, Zheng, Li, Yang, Wu, Lin, and Cai}]{chi2023normkd}
Zhihao Chi, Tu~Zheng, Hengjia Li, Zheng Yang, Boxi Wu, Binbin Lin, and Deng Cai. 2023.
\newblock Normkd: Normalized logits for knowledge distillation.
\newblock \emph{arXiv preprint arXiv:2308.00520}.

\bibitem[{Cobbe et~al.(2021)Cobbe, Kosaraju, Bavarian, Chen, Jun, Kaiser, Plappert, Tworek, Hilton, Nakano et~al.}]{cobbe2021training}
Karl Cobbe, Vineet Kosaraju, Mohammad Bavarian, Mark Chen, Heewoo Jun, Lukasz Kaiser, Matthias Plappert, Jerry Tworek, Jacob Hilton, Reiichiro Nakano, et~al. 2021.
\newblock Training verifiers to solve math word problems.
\newblock \emph{arXiv preprint arXiv:2110.14168}.

\bibitem[{Cui et~al.(2024)Cui, Yuan, Ding, Yao, He, Zhu, Ni, Xie, Xie, Lin et~al.}]{cui2024ultrafeedback}
Ganqu Cui, Lifan Yuan, Ning Ding, Guanming Yao, Bingxiang He, Wei Zhu, Yuan Ni, Guotong Xie, Ruobing Xie, Yankai Lin, et~al. 2024.
\newblock Ultrafeedback: Boosting language models with scaled ai feedback.
\newblock In \emph{Proceedings of ICML}.

\bibitem[{Deletang et~al.(2024)Deletang, Ruoss, Duquenne, Catt, Genewein, Mattern, Grau-Moya, Wenliang, Aitchison, Orseau et~al.}]{deletanglanguage}
Gregoire Deletang, Anian Ruoss, Paul-Ambroise Duquenne, Elliot Catt, Tim Genewein, Christopher Mattern, Jordi Grau-Moya, Li~Kevin Wenliang, Matthew Aitchison, Laurent Orseau, et~al. 2024.
\newblock Language modeling is compression.
\newblock In \emph{The Twelfth International Conference on Learning Representations}.

\bibitem[{Ding et~al.(2023)Ding, Chen, Xu, Qin, Hu, Liu, Sun, and Zhou}]{ding2023enhancing}
Ning Ding, Yulin Chen, Bokai Xu, Yujia Qin, Shengding Hu, Zhiyuan Liu, Maosong Sun, and Bowen Zhou. 2023.
\newblock Enhancing chat language models by scaling high-quality instructional conversations.
\newblock In \emph{Proceedings of the 2023 Conference on Empirical Methods in Natural Language Processing}, pages 3029--3051.

\bibitem[{Ding et~al.(2019)Ding, Wu, Sun, Guo, and Xia}]{ding2019adaptive}
Qianggang Ding, Sifan Wu, Hao Sun, Jiadong Guo, and Shu-Tao Xia. 2019.
\newblock Adaptive regularization of labels.
\newblock \emph{arXiv preprint arXiv:1908.05474}.

\bibitem[{Duan(2016)}]{10.1007/978-3-319-50496-4_89}
Nan Duan. 2016.
\newblock Overview of the nlpcc-iccpol 2016 shared task: Open domain chinese question answering.
\newblock In \emph{Natural Language Understanding and Intelligent Applications}, pages 942--948. Springer International Publishing.

\bibitem[{Duan and Tang(2018)}]{10.1007/978-3-319-73618-1_86}
Nan Duan and Duyu Tang. 2018.
\newblock Overview of the nlpcc 2017 shared task: Open domain chinese question answering.
\newblock In \emph{Natural Language Processing and Chinese Computing}, pages 954--961. Springer International Publishing.

\bibitem[{Dubey et~al.(2024)Dubey, Jauhri, Pandey, Kadian, Al-Dahle, Letman, Mathur, Schelten, Yang, Fan et~al.}]{dubey2024llama}
Abhimanyu Dubey, Abhinav Jauhri, Abhinav Pandey, Abhishek Kadian, Ahmad Al-Dahle, Aiesha Letman, Akhil Mathur, Alan Schelten, Amy Yang, Angela Fan, et~al. 2024.
\newblock The llama 3 herd of models.
\newblock \emph{arXiv preprint arXiv:2407.21783}.

\bibitem[{Fukuda et~al.(2017)Fukuda, Suzuki, Kurata, Thomas, Cui, and Ramabhadran}]{fukuda2017efficient}
Takashi Fukuda, Masayuki Suzuki, Gakuto Kurata, Samuel Thomas, Jia Cui, and Bhuvana Ramabhadran. 2017.
\newblock Efficient knowledge distillation from an ensemble of teachers.
\newblock In \emph{Interspeech}, pages 3697--3701.

\bibitem[{GLM et~al.(2024)GLM, Zeng, Xu, Wang, Zhang, Yin, Rojas, Feng, Zhao, Lai et~al.}]{glm2024chatglm}
Team GLM, Aohan Zeng, Bin Xu, Bowen Wang, Chenhui Zhang, Da~Yin, Diego Rojas, Guanyu Feng, Hanlin Zhao, Hanyu Lai, et~al. 2024.
\newblock Chatglm: A family of large language models from glm-130b to glm-4 all tools.
\newblock \emph{arXiv preprint arXiv:2406.12793}.

\bibitem[{Gou et~al.(2021)Gou, Yu, Maybank, and Tao}]{gou2021knowledge}
Jianping Gou, Baosheng Yu, Stephen~J Maybank, and Dacheng Tao. 2021.
\newblock Knowledge distillation: A survey.
\newblock \emph{International Journal of Computer Vision}, 129(6):1789--1819.

\bibitem[{Gu et~al.(2024)Gu, Dong, Wei, and Huang}]{gu2024minillm}
Yuxian Gu, Li~Dong, Furu Wei, and Minlie Huang. 2024.
\newblock Minillm: Knowledge distillation of large language models.
\newblock In \emph{The Twelfth International Conference on Learning Representations}.

\bibitem[{Gunter et~al.(2024)Gunter, Wang, Wang, Pang, Narayanan, Zhang, Zhang, Chen, Chiu, Qiu et~al.}]{gunter2024apple}
Tom Gunter, Zirui Wang, Chong Wang, Ruoming Pang, Andy Narayanan, Aonan Zhang, Bowen Zhang, Chen Chen, Chung-Cheng Chiu, David Qiu, et~al. 2024.
\newblock Apple intelligence foundation language models.
\newblock \emph{arXiv preprint arXiv:2407.21075}.

\bibitem[{Habib et~al.(2023)Habib, Saleem, and Lall}]{habib2023knowledge}
Gousia Habib, Tausifa~Jan Saleem, and Brejesh Lall. 2023.
\newblock Knowledge distillation in vision transformers: A critical review.
\newblock \emph{arXiv preprint arXiv:2302.02108}.

\bibitem[{Hendrycks et~al.(2021)Hendrycks, Burns, Basart, Zou, Mazeika, Song, and Steinhardt}]{hendrycksmeasuring}
Dan Hendrycks, Collin Burns, Steven Basart, Andy Zou, Mantas Mazeika, Dawn Song, and Jacob Steinhardt. 2021.
\newblock Measuring massive multitask language understanding.
\newblock In \emph{International Conference on Learning Representations}.

\bibitem[{Hinton(2015)}]{hinton2015distilling}
Geoffrey Hinton. 2015.
\newblock Distilling the knowledge in a neural network.
\newblock \emph{arXiv preprint arXiv:1503.02531}.

\bibitem[{Holtzman et~al.(2019)Holtzman, Buys, Du, Forbes, and Choi}]{holtzmancurious}
Ari Holtzman, Jan Buys, Li~Du, Maxwell Forbes, and Yejin Choi. 2019.
\newblock The curious case of neural text degeneration.
\newblock In \emph{International Conference on Learning Representations}.

\bibitem[{Hou et~al.(2020)Hou, Huang, Shang, Jiang, Chen, and Liu}]{hou2020dynabert}
Lu~Hou, Zhiqi Huang, Lifeng Shang, Xin Jiang, Xiao Chen, and Qun Liu. 2020.
\newblock Dynabert: Dynamic bert with adaptive width and depth.
\newblock \emph{Advances in Neural Information Processing Systems}, 33:9782--9793.

\bibitem[{Hu et~al.(2024)Hu, Tu, Han, He, Cui, Long, Zheng, Fang, Huang, Zhao et~al.}]{hu2024minicpm}
Shengding Hu, Yuge Tu, Xu~Han, Chaoqun He, Ganqu Cui, Xiang Long, Zhi Zheng, Yewei Fang, Yuxiang Huang, Weilin Zhao, et~al. 2024.
\newblock Minicpm: Unveiling the potential of small language models with scalable training strategies.
\newblock \emph{arXiv preprint arXiv:2404.06395}.

\bibitem[{Huang et~al.(2024)Huang, Bai, Zhu, Zhang, Zhang, Su, Liu, Lv, Zhang, Fu et~al.}]{huang2024c}
Yuzhen Huang, Yuzhuo Bai, Zhihao Zhu, Junlei Zhang, Jinghan Zhang, Tangjun Su, Junteng Liu, Chuancheng Lv, Yikai Zhang, Yao Fu, et~al. 2024.
\newblock C-eval: A multi-level multi-discipline chinese evaluation suite for foundation models.
\newblock \emph{Advances in Neural Information Processing Systems}, 36.

\bibitem[{Jiao et~al.(2020)Jiao, Yin, Shang, Jiang, Chen, Li, Wang, and Liu}]{jiao2020tinybert}
Xiaoqi Jiao, Yichun Yin, Lifeng Shang, Xin Jiang, Xiao Chen, Linlin Li, Fang Wang, and Qun Liu. 2020.
\newblock Tinybert: Distilling bert for natural language understanding.
\newblock In \emph{Findings of the Association for Computational Linguistics: EMNLP 2020}, pages 4163--4174.

\bibitem[{Kalamkar et~al.(2019)Kalamkar, Mudigere, Mellempudi, Das, Banerjee, Avancha, Vooturi, Jammalamadaka, Huang, Yuen et~al.}]{kalamkar2019study}
Dhiraj Kalamkar, Dheevatsa Mudigere, Naveen Mellempudi, Dipankar Das, Kunal Banerjee, Sasikanth Avancha, Dharma~Teja Vooturi, Nataraj Jammalamadaka, Jianyu Huang, Hector Yuen, et~al. 2019.
\newblock A study of bfloat16 for deep learning training.
\newblock \emph{arXiv preprint arXiv:1905.12322}.

\bibitem[{Kenton and Toutanova(2019)}]{kenton2019bert}
Jacob Devlin Ming-Wei~Chang Kenton and Lee~Kristina Toutanova. 2019.
\newblock Bert: Pre-training of deep bidirectional transformers for language understanding.
\newblock In \emph{Proceedings of NAACL-HLT}.

\bibitem[{Kiefel and Shah(2024)}]{kiefel2024lokilm}
Justin Kiefel and Shrey Shah. 2024.
\newblock Lokilm: Technical report.
\newblock \emph{arXiv preprint arXiv:2407.07370}.

\bibitem[{Kim and Kim(2017)}]{kim2017transferring}
Seungwook Kim and Hyo{-}Eun Kim. 2017.
\newblock \href {https://openreview.net/forum?id=ByXrfaGFe} {Transferring knowledge to smaller network with class-distance loss}.

\bibitem[{Kim et~al.(2021)Kim, Oh, Kim, Cho, and Yun}]{kimcomparing}
Taehyeon Kim, Jaehoon Oh, Nak~Yil Kim, Sangwook Cho, and Se-Young Yun. 2021.
\newblock Comparing kullback-leibler divergence and mean squared error loss in knowledge distillation.

\bibitem[{Kim and Rush(2016)}]{kim2016sequence}
Yoon Kim and Alexander~M Rush. 2016.
\newblock Sequence-level knowledge distillation.
\newblock In \emph{Proceedings of the 2016 Conference on Empirical Methods in Natural Language Processing}, pages 1317--1327.

\bibitem[{Kingma(2014)}]{kingma2014adam}
Diederik~P Kingma. 2014.
\newblock Adam: A method for stochastic optimization.
\newblock \emph{arXiv preprint arXiv:1412.6980}.

\bibitem[{Komodakis and Zagoruyko(2017)}]{komodakis2017paying}
Nikos Komodakis and Sergey Zagoruyko. 2017.
\newblock Paying more attention to attention: improving the performance of convolutional neural networks via attention transfer.
\newblock In \emph{Proceedings of ICLR}.

\bibitem[{Liu et~al.(2024)Liu, Feng, Wang, Wang, Liu, Zhao, Dengr, Ruan, Dai, Guo et~al.}]{liu2024deepseek}
Aixin Liu, Bei Feng, Bin Wang, Bingxuan Wang, Bo~Liu, Chenggang Zhao, Chengqi Dengr, Chong Ruan, Damai Dai, Daya Guo, et~al. 2024.
\newblock Deepseek-v2: A strong, economical, and efficient mixture-of-experts language model.
\newblock \emph{arXiv preprint arXiv:2405.04434}.

\bibitem[{Liu et~al.(2020)Liu, Zhou, Wang, Zhao, Deng, and Ju}]{liu2020fastbert}
Weijie Liu, Peng Zhou, Zhiruo Wang, Zhe Zhao, Haotang Deng, and Qi~Ju. 2020.
\newblock Fastbert: a self-distilling bert with adaptive inference time.
\newblock In \emph{Proceedings of ACL}, pages 6035--6044.

\bibitem[{Mirzadeh et~al.(2020)Mirzadeh, Farajtabar, Li, Levine, Matsukawa, and Ghasemzadeh}]{mirzadeh2020improved}
Seyed~Iman Mirzadeh, Mehrdad Farajtabar, Ang Li, Nir Levine, Akihiro Matsukawa, and Hassan Ghasemzadeh. 2020.
\newblock Improved knowledge distillation via teacher assistant.
\newblock In \emph{Proceedings of the AAAI conference on artificial intelligence}, volume~34, pages 5191--5198.

\bibitem[{Mitra et~al.(2023)Mitra, Del~Corro, Mahajan, Codas, Simoes, Agarwal, Chen, Razdaibiedina, Jones, Aggarwal et~al.}]{mitra2023orca}
Arindam Mitra, Luciano Del~Corro, Shweti Mahajan, Andres Codas, Clarisse Simoes, Sahaj Agarwal, Xuxi Chen, Anastasia Razdaibiedina, Erik Jones, Kriti Aggarwal, et~al. 2023.
\newblock Orca 2: Teaching small language models how to reason.
\newblock \emph{arXiv preprint arXiv:2311.11045}.

\bibitem[{M{\"u}ller et~al.(2019)M{\"u}ller, Kornblith, and Hinton}]{muller2019does}
Rafael M{\"u}ller, Simon Kornblith, and Geoffrey~E Hinton. 2019.
\newblock When does label smoothing help?
\newblock \emph{Advances in neural information processing systems}, 32.

\bibitem[{Muralidharan et~al.(2024)Muralidharan, Sreenivas, Joshi, Chochowski, Patwary, Shoeybi, Catanzaro, Kautz, and Molchanov}]{muralidharan2024compact}
Saurav Muralidharan, Sharath~Turuvekere Sreenivas, Raviraj Joshi, Marcin Chochowski, Mostofa Patwary, Mohammad Shoeybi, Bryan Catanzaro, Jan Kautz, and Pavlo Molchanov. 2024.
\newblock Compact language models via pruning and knowledge distillation.
\newblock \emph{arXiv preprint arXiv:2407.14679}.

\bibitem[{OpenAI(2022)}]{openai2022chatgpt}
OpenAI. 2022.
\newblock Chatgpt.
\newblock \url{https://openai.com/index/chatgpt/}.
\newblock Accessed: 2024-10-04.

\bibitem[{OpenAI(2023)}]{openai2023gpt}
OpenAI. 2023.
\newblock \href {https://arxiv.org/pdf/2303.08774.pdf} {{GPT}-4 technical report}.
\newblock \emph{arXiv preprint arXiv:2303.08774}.

\bibitem[{Ouyang et~al.(2022)Ouyang, Wu, Jiang, Almeida, Wainwright, Mishkin, Zhang, Agarwal, Slama, Ray et~al.}]{ouyang2022training}
Long Ouyang, Jeffrey Wu, Xu~Jiang, Diogo Almeida, Carroll Wainwright, Pamela Mishkin, Chong Zhang, Sandhini Agarwal, Katarina Slama, Alex Ray, et~al. 2022.
\newblock Training language models to follow instructions with human feedback.
\newblock \emph{Advances in neural information processing systems}, 35:27730--27744.

\bibitem[{Qi et~al.(2024)Qi, Peng, Wang, Xu, Hou, and Li}]{qi2024adelie}
Yunjia Qi, Hao Peng, Xiaozhi Wang, Bin Xu, Lei Hou, and Juanzi Li. 2024.
\newblock Adelie: Aligning large language models on information extraction.
\newblock \emph{arXiv preprint arXiv:2405.05008}.

\bibitem[{Sakaguchi et~al.(2020)Sakaguchi, Le~Bras, Bhagavatula, and Choi}]{sakaguchi2020winogrande}
Keisuke Sakaguchi, Ronan Le~Bras, Chandra Bhagavatula, and Yejin Choi. 2020.
\newblock Winogrande: An adversarial winograd schema challenge at scale.
\newblock In \emph{Proceedings of AAAI}, volume~34, pages 8732--8740.

\bibitem[{Sanh et~al.(2019)Sanh, Debut, Chaumond, and Wolf}]{sanh2019distilbert}
Victor Sanh, Lysandre Debut, Julien Chaumond, and Thomas Wolf. 2019.
\newblock Distilbert, a distilled version of bert: smaller, faster, cheaper and lighter.
\newblock In \emph{NeurIPS $EMC^2$ Workshop}.

\bibitem[{Soviany et~al.(2022)Soviany, Ionescu, Rota, and Sebe}]{soviany2022curriculum}
Petru Soviany, Radu~Tudor Ionescu, Paolo Rota, and Nicu Sebe. 2022.
\newblock Curriculum learning: A survey.
\newblock \emph{International Journal of Computer Vision}, 130(6):1526--1565.

\bibitem[{Strubell et~al.(2019)Strubell, Ganesh, and Mccallum}]{strubell2019energy}
Emma Strubell, Ananya Ganesh, and Andrew Mccallum. 2019.
\newblock Energy and policy considerations for deep learning in nlp.
\newblock In \emph{Proceedings of ACL}, pages 3645--3650.

\bibitem[{Sun et~al.(2024)Sun, Liu, Li, Wang, Dong, Lin, and Huang}]{sun2024conifer}
Haoran Sun, Lixin Liu, Junjie Li, Fengyu Wang, Baohua Dong, Ran Lin, and Ruohui Huang. 2024.
\newblock Conifer: Improving complex constrained instruction-following ability of large language models.
\newblock \emph{arXiv preprint arXiv:2404.02823}.

\bibitem[{Sun et~al.(2020{\natexlab{a}})Sun, Yu, Yu, and Cardie}]{sun2020investigating}
Kai Sun, Dian Yu, Dong Yu, and Claire Cardie. 2020{\natexlab{a}}.
\newblock Investigating prior knowledge for challenging chinese machine reading comprehension.
\newblock \emph{Transactions of the Association for Computational Linguistics}, 8:141--155.

\bibitem[{Sun et~al.(2020{\natexlab{b}})Sun, Yu, Song, Liu, Yang, and Zhou}]{sun2020mobilebert}
Zhiqing Sun, Hongkun Yu, Xiaodan Song, Renjie Liu, Yiming Yang, and Denny Zhou. 2020{\natexlab{b}}.
\newblock Mobilebert: a compact task-agnostic bert for resource-limited devices.
\newblock In \emph{Proceedings of ACL}, pages 2158--2170.

\bibitem[{Tan and Wang(2021)}]{tan2021towards}
Ke~Tan and DeLiang Wang. 2021.
\newblock Towards model compression for deep learning based speech enhancement.
\newblock \emph{IEEE/ACM transactions on audio, speech, and language processing}, 29:1785--1794.

\bibitem[{Taori et~al.(2023)Taori, Gulrajani, Zhang, Dubois, Li, Guestrin, Liang, and B.~Hashimoto}]{alpaca}
Rohan Taori, Ishaan Gulrajani, Tianyi Zhang, Yann Dubois, Xuechen Li, Carlos Guestrin, Percy Liang, and Tatsunori B.~Hashimoto. 2023.
\newblock \href {https://crfm.stanford.edu/2023/03/13/alpaca.html} {Alpaca: {A} strong, replicable instruction-following model}.

\bibitem[{Team et~al.(2024)Team, Riviere, Pathak, Sessa, Hardin, Bhupatiraju, Hussenot, Mesnard, Shahriari, Ram{\'e} et~al.}]{team2024gemma}
Gemma Team, Morgane Riviere, Shreya Pathak, Pier~Giuseppe Sessa, Cassidy Hardin, Surya Bhupatiraju, L{\'e}onard Hussenot, Thomas Mesnard, Bobak Shahriari, Alexandre Ram{\'e}, et~al. 2024.
\newblock Gemma 2: Improving open language models at a practical size.
\newblock \emph{arXiv preprint arXiv:2408.00118}.

\bibitem[{Team(2024)}]{qwen2.5}
Qwen Team. 2024.
\newblock \href {https://qwenlm.github.io/blog/qwen2.5/} {Qwen2.5: A party of foundation models}.

\bibitem[{Tian et~al.(2020)Tian, Krishnan, and Isola}]{tiancontrastive2020}
Yonglong Tian, Dilip Krishnan, and Phillip Isola. 2020.
\newblock Contrastive representation distillation.
\newblock In \emph{Proceedings of ICLR}.

\bibitem[{Turuvekere~Sreenivas et~al.(2024)Turuvekere~Sreenivas, Muralidharan, Joshi, Chochowski, Patwary, Shoeybi, Catanzaro, Kautz, and Molchanov}]{turuvekere2024llm}
Sharath Turuvekere~Sreenivas, Saurav Muralidharan, Raviraj Joshi, Marcin Chochowski, Mostofa Patwary, Mohammad Shoeybi, Bryan Catanzaro, Jan Kautz, and Pavlo Molchanov. 2024.
\newblock Llm pruning and distillation in practice: The minitron approach.
\newblock \emph{arXiv e-prints}, pages arXiv--2408.

\bibitem[{Vicuna(2023)}]{vicuna}
Vicuna. 2023.
\newblock \href {https://lmsys.org/blog/2023-03-30-vicuna/} {Vicuna: {A}n open-source chatbot impressing {GPT}-4 with 90\%* {C}hat{GPT} quality}.

\bibitem[{Wang et~al.(2020{\natexlab{a}})Wang, Wei, Dong, Bao, Yang, and Zhou}]{wang2020minilm}
Wenhui Wang, Furu Wei, Li~Dong, Hangbo Bao, Nan Yang, and Ming Zhou. 2020{\natexlab{a}}.
\newblock Minilm: Deep self-attention distillation for task-agnostic compression of pre-trained transformers.
\newblock \emph{Advances in Neural Information Processing Systems}, 33:5776--5788.

\bibitem[{Wang et~al.(2020{\natexlab{b}})Wang, Fu, Liao, Wang, Lei, and Mei}]{wang2020exclusivity}
Xiaobo Wang, Tianyu Fu, Shengcai Liao, Shuo Wang, Zhen Lei, and Tao Mei. 2020{\natexlab{b}}.
\newblock Exclusivity-consistency regularized knowledge distillation for face recognition.
\newblock In \emph{Computer Vision--ECCV 2020: 16th European Conference, Glasgow, UK, August 23--28, 2020, Proceedings, Part XXIV 16}, pages 325--342. Springer.

\bibitem[{Wei and Bai(2024)}]{wei2024dynamic}
Yukang Wei and Yu~Bai. 2024.
\newblock Dynamic temperature knowledge distillation.
\newblock \emph{arXiv preprint arXiv:2404.12711}.

\bibitem[{Xu et~al.(2023)Xu, Sun, Zheng, Geng, Zhao, Feng, Tao, and Jiang}]{xu2023wizardlm}
Can Xu, Qingfeng Sun, Kai Zheng, Xiubo Geng, Pu~Zhao, Jiazhan Feng, Chongyang Tao, and Daxin Jiang. 2023.
\newblock Wizardlm: Empowering large language models to follow complex instructions.
\newblock \emph{arXiv preprint arXiv:2304.12244}.

\bibitem[{Xu et~al.(2020)Xu, Zhou, Ge, Wei, and Zhou}]{xu2020bert}
Canwen Xu, Wangchunshu Zhou, Tao Ge, Furu Wei, and Ming Zhou. 2020.
\newblock Bert-of-theseus: Compressing bert by progressive module replacing.
\newblock In \emph{Proceedings of EMNLP}, pages 7859--7869.

\bibitem[{Xu et~al.(2024)Xu, Li, Tao, Shen, Cheng, Li, Xu, Tao, and Zhou}]{xu2024survey}
Xiaohan Xu, Ming Li, Chongyang Tao, Tao Shen, Reynold Cheng, Jinyang Li, Can Xu, Dacheng Tao, and Tianyi Zhou. 2024.
\newblock A survey on knowledge distillation of large language models.
\newblock \emph{arXiv preprint arXiv:2402.13116}.

\bibitem[{Yim et~al.(2017)Yim, Joo, Bae, and Kim}]{yim2017gift}
Junho Yim, Donggyu Joo, Jihoon Bae, and Junmo Kim. 2017.
\newblock A gift from knowledge distillation: Fast optimization, network minimization and transfer learning.
\newblock In \emph{Proceedings of the IEEE conference on computer vision and pattern recognition}, pages 4133--4141.

\bibitem[{Zellers et~al.(2019)Zellers, Holtzman, Bisk, Farhadi, and Choi}]{zellers2019hellaswag}
Rowan Zellers, Ari Holtzman, Yonatan Bisk, Ali Farhadi, and Yejin Choi. 2019.
\newblock Hellaswag: Can a machine really finish your sentence?
\newblock In \emph{Proceedings of ACL}, pages 4791--4800.

\bibitem[{Zhang et~al.(2020)Zhang, Hou, Yin, Shang, Chen, Jiang, and Liu}]{zhang2020ternarybert}
Wei Zhang, Lu~Hou, Yichun Yin, Lifeng Shang, Xiao Chen, Xin Jiang, and Qun Liu. 2020.
\newblock Ternarybert: Distillation-aware ultra-low bit bert.
\newblock In \emph{Proceedings of EMNLP}, pages 509--521.

\bibitem[{Zhao et~al.(2022)Zhao, Cui, Song, Qiu, and Liang}]{zhao2022decoupled}
Borui Zhao, Quan Cui, Renjie Song, Yiyu Qiu, and Jiajun Liang. 2022.
\newblock Decoupled knowledge distillation.
\newblock In \emph{Proceedings of the IEEE/CVF Conference on computer vision and pattern recognition}, pages 11953--11962.

\bibitem[{Zheng and YANG(2024)}]{zhengknowledge}
Kaixiang Zheng and EN-HUI YANG. 2024.
\newblock Knowledge distillation based on transformed teacher matching.
\newblock In \emph{The Twelfth International Conference on Learning Representations}.

\end{thebibliography}

\clearpage
\appendix

%\section*{Appendices}
\section{Experimental Details and more Results}

% 这一章节介绍实验的细节和更多的实验结果。所有的实验都是在H800 GPU上进行的，大约消耗43,000 GPU hours。
This section introduces the experimental details and additional results. All experiments are conducted on Nvidia H800 GPUs.

\subsection{Preliminary Experiment}
\label{sec:app_preliminary}
% 我们训练的1.9B模型的模型结构见表1。在SFT阶段，我们使用了10B的high-quality insturct-tuning data和额外10B的pre-training text corpus，对于instruct-tuning的数据，我们只计算response部分的数据。As for evaluation, for HellaSwag, WinoGrande, PIQA, KBQA, We adopt zero-shot evaluation. For C3 and C-Eval, we adopt 5-shot evaluation. For MMLU, we adopt 6-shot evaluation. For GSM8k, we adopt 8-shot evaluation. 

The architecture of the 1.9B student LLM is shown in Table~\ref{tab:app_model_archi}. For the SFT phase, we utilize a mixture of 10B high-quality instruction-tuning data and an additional 10B pre-training text corpus. For the instruction-tuning data, we only compute the language modeling loss for the response part. In the SFT stage, we adopt a $256$ batch size, a cosine learning rate scheduler with $4\times10^{-5}$ maximum learning rate, $4\times10^{-6}$ minimum learning rate, and $1\%$ warmup rate. 
As for evaluation, we adopt zero-shot evaluation for HellaSwag, WinoGrande, PIQA, and KBQA; 5-shot evaluation for C3 and C-Eval; 6-shot evaluation for MMLU; and 8-shot evaluation for GSM8k. We set the sampling temperature to $0$.

\subsection{Logits Processing}
\label{sec:app_logits}
We first employ NormKD~\citep{chi2023normkd} and WTTM~\citep{zhengknowledge} as the adaptive temperature calculation methods. Our implementation differs slightly from the original versions, as we use truncated logits instead of logits of the entire vocabulary. For NormKD, we set the hyper-parameter \texttt{T\_norm} to $1.0$ and $\alpha$ to $0.5$ in Equation~\ref{eq:eq1}. For WTTM, we set the hyper-parameters $\gamma$ to $0.1$ and $\beta$ to $1.0$. 
% We also propose two compact adaptive temperature calculation methods: AdaKD\textsubscript{SD}, which adopts the standard deviation as the temperature $\tau$; AdaKD\textsubscript{H}, which adopts $\tau_H$ in Equation~\ref{eq:entropy}.
% \begin{equation}
% \label{eq:entropy}
% \tau_H = \tau_{\text{max}} - (\tau_{\text{max}} - \tau_{\text{min}}) \times \frac{H}{H_{\text{max}}}
% \end{equation}
For $\tau_H$ in Equation~\ref{eq:entropy}, $H$ denotes the entropy of each sample, and $H_{\text{max}}$ is the largest entropy and is estimated on 10 million tokens.
We set $\tau_{\text{max}}=2.0$, $\tau_{\text{min}}=0.1$, and $H_{\text{max}}=4.8$. Experimental results of \cref{sec:logits_exp} on all evaluation datasets are presented in \cref{tab:app_pk,tab:app_temperature}.

% dynamic 温度
% NormKD~\citep{chi2023normkd} and WTTM~\citep{zhengknowledge}
% 我们使用了NormKD和WTTM作为adaptive temperature的方法。其实现和原论文不同之处在于我们没用整个词表的logits，我们用的是截断之后的logits。对于NormKD，我们设置其原论文中的超参数T_norm=1.0，设置Equation 1中的a=0.5。对于WTTM，我们设置其原论文中的超参数r=0.1, b=1。
% 我们自己设计了简洁的adaptive temperature计算的方法，AdaKD.

\subsection{Loss Selection}
\label{sec:app_loss}
For the WSD scheduler~\citep{hu2024minicpm}, we adopt a linear scheduler during the warmup stage and a cosine scheduler during the decay stage. The experimental results using different $\alpha$ on all the evaluation datasets are shown in Table~\ref{tab:app_a}.

\subsection{Model Size}
\label{sec:app_model_size}
The architectures of different sizes of student LLMs are shown in Table~\ref{tab:app_model_archi}. When pre-training 1.9B and 3.8B student LLMs on 500 billion tokens, we save a checkpoint every $10,000$ optimization step. We also save the checkpoint at the end. For each checkpoint, we conduct SFT as in the preliminary experiment before evaluation. The results on all evaluation datasets are shown in \cref{tab:app_model_size,tab:app_corpus_size}. We report the averaged performance in Figure~\ref{fig:corpus_size}.

\subsection{\textit{Offline} or \textit{Online}}
\label{sec:app_online}
% 重新预训练teacher LLM的细节
% 我们重新from scratch训练了一个9B的模型作为teacher LLM。由于很高的cost，我们只预训练了125B tokens，同时存储其logits，which消耗了64B的存储，这说明由于teacher LLM还没收敛，这些logits分布更加uniform，因此含有更多噪声。

We pre-train a new 9B LLM from scratch as the teacher LLM, with a $1,728$ batch size, $4,096$ max sequence length, a cosine learning rate scheduler with $6\times10^{-4}$ maximum learning rate, $6\times10^{-5}$ minimum learning rate, and $1\%$ warmup rate. Due to the high cost, we only adopt 400B tokens and store their logits simultaneously, which consumes about 180TB of disk storage space. This indicates that, since the teacher LLM has not yet converged, the logits are more uniform and contain more noise.

\subsection{A Better Configuration for PD}
\label{sec:app_better}
Based on our exploration, we select a better configuration for pre-training distillation. For logits processing, we use top-$0.95$-$50$ truncation and apply a temperature of $\tau=2.0$ for normalization. For loss selection, we adopt KLD as the distillation loss and combine it with LM loss using WSD-$\alpha$ and WSD-LR. The WSD hyper-parameters are the same as in \cref{sec:loss_exp}, except for the maximum value of $\alpha$, which is set to $0.9$. We use GLM-4-9B as the teacher LLM to distill 1.9B and 3.8B student LLMs. We adopt \textit{offline} logits for PD. The results on all evaluation datasets are shown in Table~\ref{tab:app_best}. We report the averaged performance in Figure~\ref{fig:figure1}.

% 最终，我们选择了一个更优的configuration for PD。For logits processing, we use top-0.95-50 for truncation and adopt a temperature t=2.0 for normalization. For loss selection, we adopt KLD as the distillation loss and use WSD-a with WSD-LR to combine it with LM loss. The WSD hyper-parameters are the same as section 1, export for the maximum of a, which we set to 0.9. We use GLM-4-9B as the teacher LLM to distill 1.9B and 3.8B teacher LLMs. We adopt the offline logits for PD. The results on all evaluation datasets are shown in Table 1. 

\begin{table*}[t]
    \centering
    \small
    \begin{adjustbox}{max width=1\linewidth}{
    \begin{tabular}{ccccccccc}
    \toprule
    & Hidden Size & FFN Hidden Size & \#Layers & \#Attention Heads & \#Query Groups & Tie \\
    \midrule
    330M & $1,024$ & $4,096$ & $12$ & $16$ & $2$ & True \\
    670M & $1,024$ & $4,096$ & $24$ & $16$ & $2$ & False \\
    1.9B & $2,048$ & $6,912$ & $24$ & $16$ & $2$ & False \\
    3.8B & $3,072$ & $8,192$ & $28$ & $24$ & $8$ & False \\
    6.8B & $4,096$ & $12,800$ & $28$ & $32$ & $8$ & False \\
    \bottomrule
    \end{tabular}
        }
\end{adjustbox}
    \caption{Model architectures of student LLMs of varying sizes. ``\#Query Groups'' denotes the number of query groups in grouped-query attention (GQA, \citealp{ainslie2023gqa}). ``Tie'' represents whether to tie the word embeddings and output weights. All the models are trained with BFLOAT16~\citep{kalamkar2019study} format.}
    \label{tab:app_model_archi}
\end{table*}

\begin{table*}[t]
    \centering
    \small
    \begin{adjustbox}{max width=1\linewidth}{
    \begin{tabular}{lrrrrrrrrr}
    \toprule
    & HellaSwag & WinoGrande & PIQA & MMLU & KBQA & C3 & C-Eval & GSM8k & Average \\
    \midrule
top-$0.5$-$100$&$54.2$&$55.8$&$72.9$&$27.1$&$3.6$&$56.3$&$28.1$&$9.8$&$38.5$\\
top-$0.6$-$100$&$55.2$&$55.0$&$73.7$&$27.2$&$2.0$&$56.6$&$25.9$&$11.0$&$38.3$\\
top-$0.7$-$100$&$54.4$&$57.5$&$72.7$&$27.8$&$2.9$&$56.7$&$27.0$&$9.4$&$38.5$\\
top-$0.8$-$100$&$54.4$&$56.7$&$72.5$&$27.0$&$3.5$&$56.0$&$26.2$&$10.6$&$38.4$\\
top-$0.85$-$100$&$54.6$&$53.7$&$73.6$&$26.2$&$3.4$&$56.5$&$26.8$&$10.8$&$38.2$\\
top-$0.9$-$100$&$53.7$&$54.9$&$72.7$&$27.9$&$3.5$&$55.5$&$28.2$&$9.2$&$38.2$\\
\midrule
top-$0.95$-$1$&$52.4$&$55.6$&$72.6$&$27.1$&$3.6$&$56.6$&$28.2$&$11.4$&$38.4$\\
top-$0.95$-$3$&$53.3$&$56.6$&$72.7$&$27.9$&$2.3$&$55.9$&$25.8$&$10.5$&$38.1$\\
top-$0.95$-$5$&$53.8$&$55.7$&$73.0$&$28.5$&$3.6$&$56.4$&$29.0$&$9.7$&$38.7$\\
top-$0.95$-$10$&$54.4$&$54.2$&$72.9$&$28.8$&$4.0$&$56.0$&$27.3$&$10.7$&$38.5$\\
top-$0.95$-$20$&$53.8$&$56.2$&$73.9$&$26.3$&$2.8$&$57.4$&$24.2$&$10.6$&$38.2$\\
top-$0.95$-$50$&$54.0$&$54.1$&$72.9$&$33.2$&$3.9$&$55.9$&$31.5$&$11.2$&$39.6$\\
\midrule
top-$0.95$-$100$&$54.2$&$55.2$&$72.5$&$27.8$&$3.5$&$55.8$&$26.7$&$10.8$&$38.3$\\
    \bottomrule
    \end{tabular}
    }
\end{adjustbox}
    \caption{Experimental results on all the evaluation datasets using different $p$ and $k$ in top-$p$-$k$ truncation.}
    \label{tab:app_pk}
\end{table*}

\begin{table*}[t]
    \centering
    \small
    \begin{adjustbox}{max width=1\linewidth}{
    \begin{tabular}{lrrrrrrrrr}
    \toprule
    & HellaSwag & WinoGrande & PIQA & MMLU & KBQA & C3 & C-Eval & GSM8k & Average \\
    \midrule
    $\tau=0.05$&$53.1$&$57.0$&$72.0$&$29.2$&$3.4$&$55.8$&$26.8$&$9.2$&$38.3$\\
$\tau=0.1$&$52.6$&$54.2$&$72.6$&$28.6$&$2.6$&$56.1$&$30.6$&$10.8$&$38.5$\\
$\tau=0.2$&$53.5$&$56.9$&$73.2$&$27.8$&$3.6$&$56.2$&$27.3$&$10.8$&$38.7$\\
$\tau=0.5$&$54.7$&$57.0$&$74.2$&$28.2$&$3.9$&$56.1$&$26.0$&$9.8$&$38.7$\\
$\tau=1.0$&$54.2$&$55.2$&$72.5$&$27.8$&$3.5$&$55.8$&$26.7$&$10.8$&$38.3$\\
$\tau=2.0$&$54.1$&$56.7$&$73.2$&$27.8$&$3.7$&$56.2$&$27.0$&$10.5$&$38.7$\\
$\tau=5.0$&$52.5$&$55.8$&$72.8$&$23.5$&$3.3$&$56.2$&$27.9$&$9.6$&$37.7$\\
$\tau=10.0$&$52.1$&$57.1$&$73.0$&$27.3$&$3.3$&$53.9$&$30.2$&$8.0$&$38.1$\\
    \bottomrule
    \end{tabular}
    }
\end{adjustbox}
    \caption{Experimental results on all the evaluation datasets using different $\tau$ in logits normalization.}
    \label{tab:app_temperature}
\end{table*}

\begin{table*}[t]
    \centering
    \small
    \begin{adjustbox}{max width=1\linewidth}{
    \begin{tabular}{lrrrrrrrrr}
    \toprule
    & HellaSwag & WinoGrande & PIQA & MMLU & KBQA & C3 & C-Eval & GSM8k & Average \\
    \midrule
$\alpha=0$&$53.3$&$54.8$&$72.9$&$28.0$&$3.6$&$54.7$&$25.9$&$8.6$&$37.7$\\
$\alpha=0.1$&$53.4$&$56.0$&$72.9$&$26.4$&$3.2$&$55.8$&$24.1$&$9.6$&$37.7$\\
$\alpha=0.5$&$53.8$&$54.4$&$72.6$&$26.9$&$3.4$&$55.9$&$29.8$&$9.6$&$38.3$\\
$\alpha=0.6$&$53.7$&$55.7$&$73.4$&$27.8$&$3.4$&$54.4$&$28.8$&$8.6$&$38.3$\\
$\alpha=0.7$&$53.6$&$56.6$&$73.4$&$28.5$&$3.8$&$55.0$&$29.6$&$10.1$&$38.8$\\
$\alpha=0.8$&$54.3$&$56.6$&$72.4$&$28.2$&$3.8$&$55.5$&$26.6$&$10.5$&$38.5$\\
$\alpha=0.9$&$55.1$&$57.4$&$73.0$&$29.6$&$3.5$&$57.2$&$25.6$&$11.1$&$39.1$\\
$\alpha=0.95$&$53.4$&$57.1$&$72.1$&$28.7$&$3.4$&$56.4$&$28.4$&$9.7$&$38.7$\\
$\alpha=1.0$&$54.2$&$55.2$&$72.5$&$27.8$&$3.5$&$55.8$&$26.7$&$10.8$&$38.3$\\
    \bottomrule
    \end{tabular}
    }
\end{adjustbox}
    \caption{Experimental results on all the evaluation datasets using different $\alpha$ in Equation~\ref{eq:eq1}.}
    \label{tab:app_a}
\end{table*}

\begin{table*}[t]
    \centering
    \small
    \begin{adjustbox}{max width=1\linewidth}{
    \begin{tabular}{lrrrrrrrrr}
    \toprule
    & HellaSwag & WinoGrande & PIQA & MMLU & KBQA & C3 & C-Eval & GSM8k & Average \\
    \midrule
    \multicolumn{9}{c}{Baseline: LM Loss} \\
    \midrule
330M&$37.4$&$54.1$&$67.4$&$24.0$&$2.0$&$47.3$&$26.2$&$2.3$&$32.6$\\
670M&$42.3$&$51.9$&$68.6$&$26.7$&$2.3$&$48.9$&$24.8$&$3.0$&$33.6$\\
1.9B&$53.3$&$54.8$&$72.9$&$28.0$&$3.6$&$54.7$&$25.9$&$8.6$&$37.7$\\
3.8B&$59.0$&$57.8$&$75.4$&$34.5$&$4.6$&$57.8$&$33.4$&$13.7$&$42.0$\\
6.8B&$63.0$&$59.9$&$75.5$&$36.7$&$4.6$&$61.8$&$37.1$&$20.9$&$44.9$\\
    \midrule
    \multicolumn{9}{c}{Teacher LLM: GLM-4-9B} \\
    \midrule
330M&$37.7$&$51.8$&$68.8$&$23.5$&$1.8$&$45.8$&$25.2$&$2.1$&$32.1$\\
670M&$43.4$&$50.9$&$69.4$&$25.7$&$2.4$&$49.4$&$26.2$&$3.1$&$33.8$\\
1.9B&$54.2$&$55.2$&$72.5$&$27.8$&$3.6$&$55.8$&$26.7$&$10.8$&$38.3$\\
3.8B&$61.4$&$60.2$&$75.6$&$39.1$&$5.0$&$61.0$&$39.5$&$17.1$&$44.9$\\
6.8B&$66.0$&$62.3$&$76.3$&$41.2$&$5.7$&$64.4$&$43.0$&$25.5$&$48.0$\\
    \midrule
    \multicolumn{9}{c}{Teacher LLM: GLM-4-32B} \\
    \midrule
330M&$37.1$&$51.5$&$67.4$&$24.2$&$2.0$&$45.2$&$24.5$&$1.4$&$31.6$\\
670M&$43.0$&$51.5$&$69.5$&$27.0$&$2.2$&$50.2$&$26.4$&$3.9$&$34.2$\\
1.9B&$53.7$&$57.9$&$73.4$&$26.2$&$3.4$&$54.6$&$26.3$&$8.0$&$37.9$\\
3.8B&$60.8$&$57.6$&$75.0$&$33.9$&$2.7$&$60.8$&$38.0$&$14.7$&$42.9$\\
6.8B&$66.2$&$62.3$&$76.6$&$41.4$&$5.1$&$63.7$&$41.4$&$22.7$&$47.4$\\
    \bottomrule
    \end{tabular}
    }
\end{adjustbox}
    \caption{Experimental results on all the evaluation datasets of baseline LLMs trained with only LM loss and distilled LLMs using varying sizes of teacher and student LLMs.}
    \label{tab:app_model_size}
\end{table*}

\begin{table*}[t]
    \centering
    \small
    \begin{adjustbox}{max width=1\linewidth}{
    \begin{tabular}{lrrrrrrrrr}
    \toprule
    & HellaSwag & WinoGrande & PIQA & MMLU & KBQA & C3 & C-Eval & GSM8k & Average \\
  \midrule
    \multicolumn{9}{c}{1.9B LLM pre-trained with LM Loss} \\
    \midrule
10,000&$52.3$&$55.4$&$72.1$&$27.8$&$3.4$&$56.3$&$26.4$&$8.0$&$37.7$\\
20,000&$56.4$&$57.6$&$74.0$&$31.9$&$4.0$&$58.2$&$31.2$&$10.3$&$40.5$\\
30,000&$58.5$&$58.6$&$74.5$&$33.6$&$4.2$&$59.4$&$38.0$&$12.3$&$42.4$\\
40,000&$59.8$&$57.6$&$74.8$&$35.7$&$4.3$&$60.4$&$36.9$&$14.5$&$43.0$\\
50,000&$60.6$&$58.0$&$75.8$&$37.8$&$4.6$&$62.0$&$40.3$&$14.9$&$44.2$\\
59,604&$61.1$&$58.8$&$75.4$&$37.7$&$4.5$&$60.9$&$39.7$&$15.7$&$44.2$\\
  \midrule
    \multicolumn{9}{c}{1.9B LLM pre-trained with KD Loss} \\
    \midrule
10,000&$53.8$&$57.1$&$73.0$&$26.0$&$3.1$&$56.3$&$25.9$&$10.7$&$38.2$\\
20,000&$58.1$&$58.7$&$74.3$&$31.4$&$3.7$&$59.6$&$31.5$&$14.5$&$41.5$\\
30,000&$60.0$&$59.1$&$74.6$&$34.4$&$4.6$&$60.0$&$35.8$&$18.0$&$43.3$\\
40,000&$60.9$&$60.0$&$74.9$&$35.1$&$4.9$&$61.7$&$38.0$&$19.0$&$44.3$\\
50,000&$61.8$&$59.9$&$75.4$&$38.5$&$4.3$&$61.9$&$41.4$&$20.6$&$45.5$\\
59,604&$61.9$&$60.3$&$75.5$&$38.9$&$4.6$&$61.8$&$40.3$&$19.4$&$45.4$\\
  \midrule
    \multicolumn{9}{c}{3.8B LLM pre-trained with LM Loss} \\
    \midrule
10,000&$58.6$&$59.9$&$74.4$&$33.1$&$4.7$&$60.2$&$36.8$&$12.8$&$42.6$\\
20,000&$63.5$&$61.3$&$75.6$&$41.0$&$4.4$&$63.2$&$42.3$&$20.5$&$46.5$\\
30,000&$65.7$&$63.6$&$76.1$&$42.8$&$2.8$&$65.1$&$47.3$&$23.7$&$48.4$\\
40,000&$67.1$&$63.2$&$76.6$&$45.2$&$1.3$&$65.8$&$46.1$&$25.8$&$48.9$\\
50,000&$68.0$&$64.2$&$76.7$&$46.0$&$4.5$&$66.9$&$48.0$&$28.5$&$50.3$\\
59,604&$68.3$&$63.1$&$77.3$&$46.9$&$2.3$&$66.7$&$47.8$&$29.3$&$50.2$\\
      \midrule
    \multicolumn{9}{c}{3.8B LLM pre-trained with KD Loss} \\
    \midrule
10,000&$60.8$&$61.5$&$75.6$&$31.7$&$4.8$&$61.0$&$36.6$&$19.0$&$43.9$\\
20,000&$65.3$&$63.1$&$76.3$&$41.6$&$5.7$&$64.0$&$44.8$&$26.5$&$48.4$\\
30,000&$67.2$&$65.2$&$76.4$&$47.0$&$6.2$&$66.4$&$47.5$&$30.9$&$50.9$\\
40,000&$68.3$&$65.4$&$76.7$&$49.4$&$6.9$&$67.1$&$50.2$&$35.0$&$52.4$\\
50,000&$69.1$&$67.4$&$77.3$&$51.3$&$6.7$&$68.5$&$50.9$&$36.5$&$53.5$\\
59,604&$69.5$&$66.5$&$77.7$&$52.4$&$6.8$&$68.5$&$52.3$&$36.2$&$53.7$\\
    \bottomrule
    \end{tabular}
    }
\end{adjustbox}
    \caption{Experimental results on all the evaluation datasets of different checkpoints saved every $10,000$ optimization step when pre-training the LLMs on 500 billion tokens. ``59604'' is the last checkpoint saved at the end.}
    \label{tab:app_corpus_size}
\end{table*}

\begin{table*}[t]
    \centering
    \small
    \begin{adjustbox}{max width=1\linewidth}{
    \begin{tabular}{lrrrrrrrrr}
    \toprule
    & HellaSwag & WinoGrande & PIQA & MMLU & KBQA & C3 & C-Eval & GSM8k & Average \\
    \midrule
1.9B&$56.9$&$59.1$&$73.9$&$29.8$&$3.7$&$59.0$&$35.2$&$12.4$&$41.2$\\
3.8B&$62.4$&$61.2$&$76.0$&$38.1$&$5.0$&$62.8$&$38.5$&$21.5$&$45.7$\\
6.8B&$67.4$&$65.1$&$76.6$&$44.3$&$5.6$&$67.1$&$44.7$&$27.4$&$49.8$\\
    \bottomrule
    \end{tabular}
    }
\end{adjustbox}
    \caption{Experimental results on all the evaluation datasets of a better pre-training distillation configuration.}
    \label{tab:app_best}
\end{table*}

\end{document}